%% file: main.tex
\documentclass[10pt,twocolumn,letterpaper]{article}

\usepackage[pagenumbers]{cvpr} %

\input{packages}

\input{macros}

\input{info}

\begin{document}

\input{fig_text/teaser}

\input{sections/0_abstract}

\input{sections/1_intro}

\input{sections/2_rw}

\input{sections/3_method}
\input{sections/4_exp}

\input{sections/5_conclusion}

\input{sections/6_ack}

{
    \small
    \bibliographystyle{ieee_fullname}
    \bibliography{main}
}

\clearpage
\renewcommand\thefigure{S\arabic{figure}}
\setcounter{figure}{0}
\renewcommand\thetable{S\arabic{table}}
\setcounter{table}{0}
\renewcommand\theequation{S\arabic{equation}}
\setcounter{equation}{0}
\renewcommand\thealgorithm{S\arabic{algorithm}}
\setcounter{algorithm}{0}
\pagenumbering{arabic}%
\renewcommand*{\thepage}{S\arabic{page}}
\setcounter{footnote}{0}
\setcounter{page}{1}
\maketitlesupplementary
\appendix

\input{sections_supp/0_overview}
\input{sections_supp/1_tech_details}

\input{sections_supp/2_dataset_details}
\input{sections_supp/3_exp_details}

\input{sections_supp/4_more_results}

\input{sections_supp/5_ablation}

\end{document}

%% file: packages.tex
\usepackage{amsmath,amsfonts,amsthm,amssymb}
\usepackage{mathtools}
\usepackage{bm}
\usepackage{bbm}
\usepackage{nicefrac}
\usepackage{microtype}
\usepackage{lipsum}

\usepackage{gensymb}
\usepackage{color,xcolor}
\usepackage{epsfig}
\usepackage{graphicx}

\usepackage{adjustbox}
\usepackage{array}
\usepackage{booktabs}
\usepackage{colortbl}
\usepackage{wrapfig}
\usepackage{hhline}
\usepackage{multirow}
\usepackage{subcaption}
\usepackage[size=small]{caption}
\usepackage{float}
\usepackage{stfloats}

\usepackage{changepage}
\usepackage{extramarks}
\usepackage{fancyhdr}
\usepackage{lastpage}
\usepackage{setspace}
\usepackage{soul}
\usepackage{xspace}

\usepackage{url}

\usepackage{algorithm}
\usepackage{algorithmicx}
\usepackage{algpseudocode}
\usepackage{enumerate}
\usepackage{enumitem}  %
\usepackage{makecell}
\usepackage{pifont}
\usepackage{titlecaps}
\usepackage[accsupp]{axessibility}
\usepackage{framed}

\usepackage{comment}
\usepackage{textcomp}

\usepackage[pagebackref,breaklinks,colorlinks,citecolor=cvprblue]{hyperref}
\usepackage[capitalize]{cleveref}

%% file: macros.tex
\newcommand{\model}{FluidNexus\xspace}

\newcommand{\newdatasetbg}{\model-Smoke\xspace}
\newcommand{\newdatasetball}{\model-Ball\xspace}

\newcommand{\myparagraph}[1]{\vspace{0.1cm}\noindent\textbf{#1}}

\renewcommand{\paragraph}[1]{\vspace{0.1cm}\noindent\textbf{#1}}

\definecolor{cvprblue}{rgb}{0.21,0.49,0.74}

\definecolor{MyDarkBlue}{rgb}{0,0.08,1}
\definecolor{MyAqua}{rgb}{0,0.7,0.7}
\definecolor{MyDarkGreen}{rgb}{0.02,0.6,0.02}
\definecolor{MyDarkRed}{rgb}{0.8,0.02,0.02}
\definecolor{MyDarkOrange}{rgb}{0.40,0.2,0.02}
\definecolor{MyPurple}{RGB}{111,0,255}
\definecolor{MyRed}{rgb}{1.0,0.0,0.0}
\definecolor{MyGold}{rgb}{0.75,0.6,0.12}
\definecolor{MyDarkgray}{rgb}{0.66, 0.66, 0.66}

\definecolor{Cardinal}{rgb}{0.549,0.082,0.082}

%% file: info.tex
\def\paperID{1514} %
\def\confName{CVPR}
\def\confYear{2025}

\title{\model: 3D Fluid Reconstruction and Prediction from a Single Video}

\author{ Yue Gao$^{*1,2}$ \quad Hong-Xing Yu$^{*1}$ \quad Bo Zhu$^{3}$ \quad Jiajun Wu$^{1}$ \vspace{4pt} \\
    $^1$Stanford University \quad $^2$Microsoft \quad  $^3$Georgia Institute of Technology %
}

%% file: fig_text/teaser.tex
\twocolumn[{%
\renewcommand\twocolumn[1][]{#1}%
\maketitle
\vspace{-12mm}
\begin{center}
    \centering
    \captionsetup{type=figure}
    \includegraphics[width=1\textwidth]{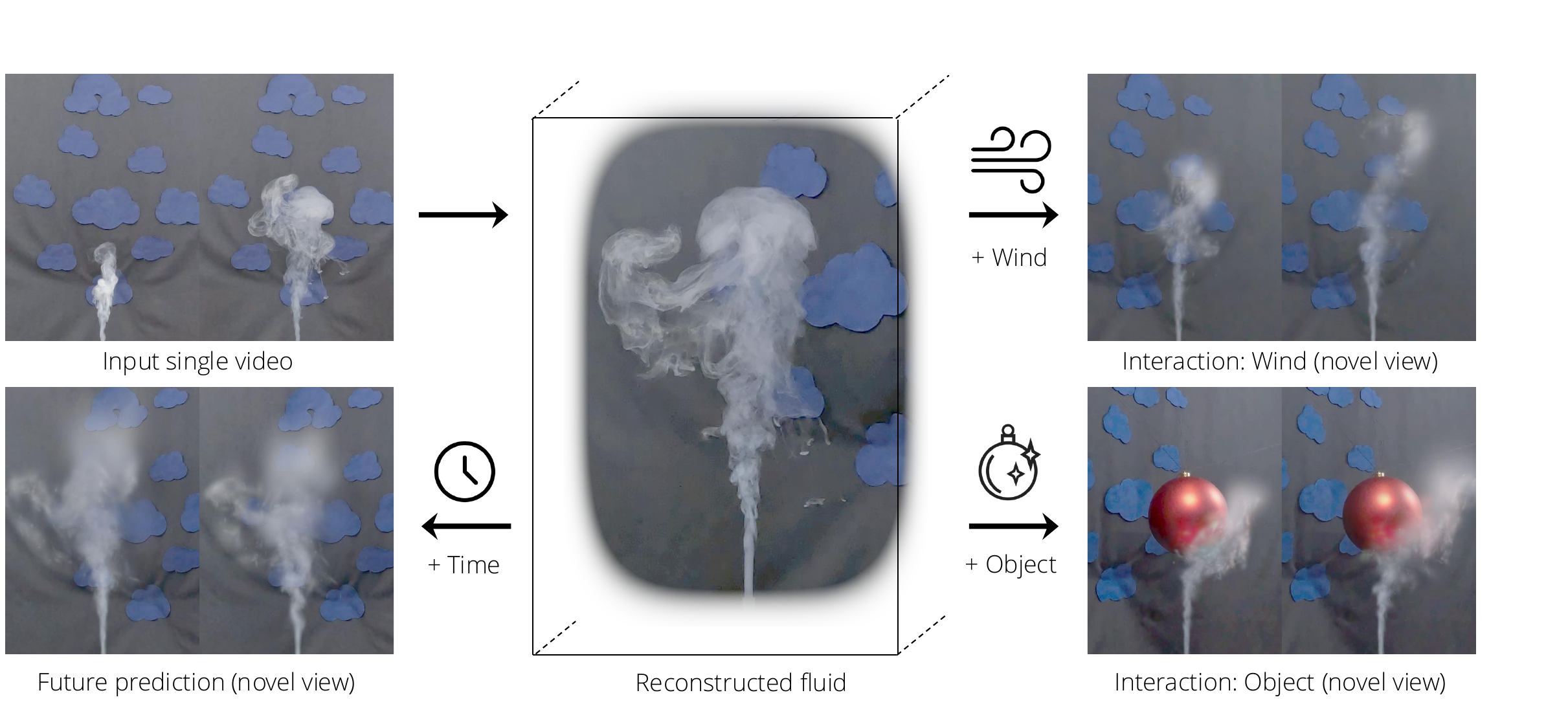}
    \vspace{-8mm}
    \caption{From a single fluid video (top left), we reconstruct the 3D fluid (middle), predict future evolution (bottom left), and simulate wind-fluid interaction (top right) / object-fluid interactions (bottom right).}
    \label{fig:teaser}
\end{center}%
}]

%% file: sections/0_abstract.tex
\begin{abstract}

\renewcommand{\thefootnote}{\fnsymbol{footnote}}\footnotetext[1]{Equal contribution.}\renewcommand{\thefootnote}{\arabic{footnote}}\setcounter{footnote}{0}

We study reconstructing and predicting 3D fluid appearance and velocity from a single video. Current methods require multi-view videos for fluid reconstruction. We present \model, a novel framework that bridges video generation and physics simulation to tackle this task. Our key insight is to synthesize multiple novel-view videos as references for reconstruction. \model consists of two key components: (1) a novel-view video synthesizer that combines frame-wise view synthesis with video diffusion refinement for generating realistic videos, and (2) a physics-integrated particle representation coupling differentiable simulation and rendering to simultaneously facilitate 3D fluid reconstruction and prediction. To evaluate our approach, we collect two new real-world fluid datasets featuring textured backgrounds and object interactions. Our method enables dynamic novel view synthesis, future prediction, and interaction simulation from a single fluid video. Project website: \url{https://yuegao.me/FluidNexus}.

\end{abstract}

%% file: sections/1_intro.tex
\vspace{-0.1cm}
\section{Introduction}

Video-based fluid reconstruction and prediction presents a promising direction for understanding fluid dynamics in scenarios where complex measurement equipment and computational fluid dynamics (CFD) simulations are impractical. This capability has broad applications in industrial monitoring~\citep{duan2013synchronized}, visual special effects~\citep{okabe2011creating}, and scientific visualization~\citep{raissi2020hidden}. Recent state-of-the-art approaches have explored integrating neural rendering with physics priors for fluid reconstruction from videos~\citep{chu2022physics,yu2024inferring}. However, these methods all require multi-view videos for reconstruction, which are often difficult to obtain in real-world scenarios. 
\vspace{1mm}

In this paper, we address a novel problem setup: \textbf{single-video 3D fluid reconstruction and prediction}. Specifically, we aim to start with a single-view video and generate multiple synchronized novel-view videos to serve as references for 3D fluid reconstruction. At first glance, this problem appears ill-posed, as a single video provides limited information about the fluid's intricate 3D structure and dynamics, while infinitely many potential 3D fluid states could correspond to the observed frames. We identify three key challenges in tackling this problem: 
(1) \textit{Single-view video to multi-view video:} From a single-view video input, we aim to synthesize realistic videos of the same scene from novel viewpoints. This task represents a conditional video-to-video translation problem with substantial viewpoint changes, which remains an open challenge for existing video generation methods.  
(2) \textit{Multi-view video to 4D reconstruction:} Using the generated multi-view videos, which may exhibit inconsistencies due to conflicts between synthesized views, we aim to reconstruct spatiotemporally consistent and physically plausible fluid flow motion. 
(3) \textit{Reconstruction to prediction:} Building upon the reconstructed 4D fluid flow data, we seek to integrate physical models to predict future fluid motion. This reconstruction-to-prediction problem is particularly challenging due to the difficulty in identifying the inherent physical priors from the reconstructed data and enforcing these constraints in future prediction.   

\vspace{1mm}
We propose a novel video-to-prediction framework \mbox{\textbf{\textit{\model}}} to address these challenges. Our system consists of two components: (1) \textit{A novel-view video synthesizer}, comprising a frame-wise novel-view diffusion model and a video diffusion refiner, which generates multi-view videos from a single-view input. The frame-wise model synthesizes individual frames from novel viewpoints without accounting for fluid dynamics, while the video diffusion refiner enhances these frames into a coherent and realistic video sequence. (2) \textit{A visual-physical particle representation}, which bridges differentiable physics and differentiable rendering using two sets of moving particles, reconstructs 4D fluid flow motion from multi-view video input. Specifically, our visual-physical particle model integrates a group of \textbf{physical particles}, which implements a differentiable physics simulator based on Position-Based Fluid (PBF)~\citep{macklin2013position}, with a group of \textbf{visual particles}, which establishes a differentiable rendering pipeline leveraging 3D Gaussian Splatting~\citep{kerbl20233d} to link the input videos with the 4D reconstructions. The interaction between these two particle groups enables a sparse yet effective representation of the spatiotemporal fluid flow during reconstruction based on multi-view video input, while simultaneously enforcing physical constraints on fluid motions during prediction, where ground truth video reference is no longer available.

\vspace{1mm}

As shown in Figure~\ref{fig:teaser}, our \model framework simultaneously addresses the multifaceted challenges of video synthesis, 4D fluid reconstruction, and future motion prediction, all starting from a single video input. We evaluated our framework on existing benchmarks as well as two newly introduced fluid datasets, which feature more complex fluid motion dynamics and intricate environmental interactions. Our key contributions can be summarized as follows:
\begin{itemize}
    \item A novel framework, \model, which enables single-video fluid reconstruction and prediction by bridging generative models and physics simulation.
    \item A reconstruction algorithm and a prediction algorithm that incorporate physics principles of 3D fluid dynamics and generative priors learned from fluid videos via differentiable simulation and rendering.
    \item Two new real fluid datasets featuring turbulent flow, textured background, and solid obstacles, which we use to train \model, and demonstrate novel view synthesis and future prediction capabilities from a single video.
\end{itemize}

%% file: sections/2_rw.tex
\section{Related Work}
\myparagraph{Video-based fluid analysis.}
Scientists and engineers have extensively studied fluid flow analysis through visible light measurements, primarily in controlled laboratories. These studies employ active sensing approaches (\eg, laser scanners~\citep{hawkins2005acquisition} and structural light~\citep{gu2012compressive}) or passive techniques like particle imaging velocimetry (PIV)~\citep{adrian2011particle,elsinga2006tomographic}. To allow broader applicability, recent methods have explored video-based fluid analysis with tomography~\citep{gregson2014capture,okabe2015fluid,zang2020tomofluid} or neural/differentiable rendering~\citep{franz2021global}. They use multi-view videos as reference, often aided with differentiable physics simulation~\citep{eckert2019scalarflow,franz2021global} to constrain the solution within a domain, \eg, GlobTrans~\citep{franz2021global} uses differentiable simulation with visual hull to constrain the reconstructed fluid. Recent methods~\citep{wang2024physics} such as Physics-Informed Neural Fluid~\citep{chu2022physics} and HyFluid~\citep{yu2024inferring} resort to physics-informed losses to help maintain physical plausibility. These methods require multiple synchronized videos as input, which are often unavailable. To address this limitation, we incorporate video generation to tackle single-video 3D fluid reconstruction and prediction.

\myparagraph{Dynamic 3D reconstruction.}
Densely reconstructing dynamic 3D scenes has been challenging~\citep{bansal20204d}. A recent breakthrough comes from representing scenes with dynamic radiance fields such as dynamic NeRFs~\citep{nerf,li2021neural,park2021hypernerf,pumarola2021d,fridovich2023k,gao2021dynamic} and dynamic 3D Gaussians~\citep{lin2024gaussian,luiten2024dynamic,li2024spacetime} and using differentiable (neural) rendering to optimize them from observed multi-view videos. By leveraging monocular depth estimation~\citep{ranftl2022midas,ke2023repurposing}, latest methods even allow reconstructing dynamic scenes from monocular videos~\citep{som2024,lei2024mosca,zhang2024monst3r}. Our approach also represents the fluid appearance with radiance fields. Different from these methods which do not consider physics modeling and thus unable to do prediction, our approach allows future prediction and interaction simulation.

\myparagraph{Physics-based dynamic scene modeling.}
Traditional fluid simulation methods span a broad spectrum, from high-fidelity computational fluid dynamics (CFD)~\citep{anderson1995computational} to real-time approximations like stable fluids~\citep{stam2023stable,selle2008unconditionally} and position-based dynamics~\citep{muller2007position,macklin2013position}. While these methods excel at physical accuracy, they often struggle with realistic appearance and detail synthesis. Recent learning-based approaches~\citep{kim2019deep} have demonstrated the ability to enhance simulation with learned fluid details~\citep{chu2017data} and dynamics~\citep{wiewel2019latent}. In parallel, computer vision community has developed physics-integrated representations that combine physical constraints with differentiable rendering~\citep{xie2024physgaussian,li2023pac,feng2024pie,feng2024gaussian}, enabling motion synthesis and interaction of objects of different materials~\citep{zhang2024physdreamer,zhong2024reconstruction}. 
However, they typically require multi-view inputs or pre-reconstructed 3D scenes. Our work bridges this gap by integrating video generation, allowing single-video fluid reconstruction and prediction.

\myparagraph{Video generation.} In the past few years, video generation models~\citep{makeavideo2022, imagenvideo2022}, especially video diffusion models~\citep{hong2022cogvideo,alignyourlatents2023,svdblattmann2023,emuvideo2023,lumiere2024,sora2024}, have been rapidly developed. Recent video generation methods, such as Sora~\citep{sora2024}, have demonstrated great promise in simulating real-world complex physical events, including fluid dynamics and interaction. Yet, the generation is in 2D, and the controllability is limited. Nevertheless, the ability to simulate realistic fluid dynamics and fluid-object interaction motivates us to integrate video diffusion models into fluid reconstruction and prediction.

%% file: sections/3_method.tex
\section{Approach}

\input{fig_text/framework}

\paragraph{Problem statement.}
Given a fixed-viewpoint video $\mathcal{V}^0=(I_1^0, I_2^0, \cdots, I_T^0)$ containing $T$ frames $I_t^0\in \mathbb{R}^{H \times W \times 3}$ of fluid dynamics (\eg, a rising plume), we aim to reconstruct 3D fluid velocity and appearance over time $T$ and predict future states beyond $T$. 

\paragraph{Our solution.}
We propose \model, which consists of a novel-view video synthesizer (left of Figure~\ref{fig:framework}) to create temporally consistent multi-view videos based on a single video input and a physics-integrated particle representation (right of Figure~\ref{fig:framework}) to reconstruct and predict fluid motion based on multi-view video inputs. We elaborate these two components in the following.

\subsection{Novel-view Video Synthesizer}
Given an input video $\mathcal{V}^0=(I_1^0, I_2^0, \cdots, I_T^0)$, we aim to generate $C$ novel-view videos $\{\mathcal{V}^c\}_{c=1}^{C}$, where $\mathcal{V}^c=(I_1^c, I_2^c, \cdots, I_T^c)$. The key challenge lies in ensuring both spatial consistency across views and temporal coherence within each view. To address this, we design our video synthesizer with a frame-wise view synthesis model and a video refinement diffusion model.

\paragraph{Frame-wise Novel View Synthesis.} To learn spatial coherence, we employ a camera view-conditioned image diffusion model to synthesize frames at novel viewpoints~\citep{zero123}. Given a single frame $I_t^0$ at timestep $t$ and a camera transform matrix $\pi_{c} \in \mathbb{R}^{3 \times 4}$ from the input view to the target view $c$, the diffusion model learns to synthesize the novel view:
\begin{equation}
\hat{I}_t^c = g(I_t^0, \pi_{c}),
\end{equation}
where $\hat{I}_t^c$ denotes the synthesized frame and $g$ denotes the diffusion model. The diffusion model performs denoising steps conditioned on the input frame $I_t^0$ and the camera transform matrix $\pi_{c}$ to generate geometrically consistent novel views. Over $t=1$ to $T$, this gives a rough video $\hat{\mathcal{V}}^c=(\hat{I}_1^c, \hat{I}_2^c, \cdots, \hat{I}_T^c)$ for a given viewpoint $c$.

\paragraph{Generative Video Refinement.}
Since each frame is generated independently of other frames, the generated frames lack temporal consistency. To address this, we introduce a generative video refinement approach conditioned on the synthesized frames to generate temporally coherent fluid videos. Our refinement extends an image editing technique, SDEdit~\cite{meng2021sdedit}, to video diffusion:
\begin{equation}\label{eqn:refinement}
(I_1^c, I_2^c, \cdots, I_T^c) = v(\hat{\mathcal{V}}^c|\lambda_\text{SDEdit}),
\end{equation}
where $v$ represents the video diffusion model and $\lambda_\text{SDEdit}$ denotes the strength of generative refinement. The refinement works as follows: intuitively, instead of denoising a sampled pure noise for $L$ steps to generate a video from scratch, it first creates an intermediate perturbed video $\widetilde{\mathcal{V}}^c$ by injecting mild noise to all frames in $\hat{\mathcal{V}}^c$, and then performs the latter $\lambda_{\text{SDEdit}}L$ denoising steps ($0<\lambda_{\text{SDEdit}}<1$). This generates a fluid video $\mathcal{V}^c$ that maintains spatial content in $\hat{\mathcal{V}}^c$ while ensuring temporal coherence by pulling it to the video manifold learned by the video diffusion model~\cite{meng2021sdedit}. The strength of this generative refinement $\lambda_{\text{SDEdit}}$ controls the balance between content preservation and temporal consistency. We leave further technical details and training details in the supplementary material.

\paragraph{Long Video Generation.}
Video diffusion models typically operate on a limited time window $T'$ smaller than our target video length $T$ due to computing limitations. Naively applying the video refinement to consecutive $T'$-frame segments leads to periodic jittering at segment boundaries. To address this, we train two video refiners: 
an unconditional $v_\text{uncond}((\hat{I}_1^c,\cdots,\hat{I}_{T'}^c))$ for generating the first segment $(I^c_1,\cdots,I^c_{T'})$;
And a $v_\text{cond}((\hat{I}_{T'+1}^c, \cdots, \hat{I}_{2T'-S}^c)|(I^c_{T'-S+1},\cdots,I^c_{T'}))$ conditioned on a few $S$ fixed previous frames for extending an existing segment. We recursively use the conditioned refiner $v_\text{cond}$ to progressively extend the sequence by $T'-S$ frames at a time, maintaining consistency with previously generated frames.

\subsection{Visual-Physical Particle Fluid Representation}
While the generated multi-view videos $\{\mathcal{V}^c\}_{c=1}^C$ provide spatial-temporal references, they lack physical plausibility in fluid dynamics. To address this, our representation integrates Position-Based Fluid (PBF)~\citep{macklin2013position} simulation, known for its efficiency and flexibility, with 3D Gaussian Splatting~\citep{kerbl20233d} to bridge simulated particles with the generated videos.

\paragraph{Particle-based Fluid Representation.}
Our fluid representation consists of two types of particles: physical particles that represent the fluid velocity field and density field in ambient 3D space, and visual particles that represent the fluid appearance and are passively advected by the velocity field. We need two different types of particles because intrinsic difference between the distributions: the velocity is defined in ambient 3D space while the appearance (visible fluid) is generally defined only in specific regions.  

Physical particles at any timestep $t$ are defined by their positions $\mathbf{p}_t\in\mathbb{R}^{3N_t^\text{physical}}$ where $N_t^\text{physical}$ denotes the physical particle count at $t$, and the associated particle velocities $\mathbf{u}_t\in\mathbb{R}^{3N_t^\text{physical}}$ (not to be confused with the velocity field $\mathbf{V}_t: \mathbb{R}^3\mapsto\mathbb{R}^3$). The velocity at any 3D point $\mathbf{x}$ at timestep $t$ is computed through kernel-weighted interpolation:
\begin{equation}
\mathbf{V}_t(\mathbf{x}|\mathbf{u}_t,\mathbf{p}_t)=\frac{\sum_j \mathbf{u}_{t,j} K(\mathbf{x}-\mathbf{p}_{t,j})}{\sum_j K(\mathbf{x}-\mathbf{p}_{t,j})},
\end{equation}
where $K(\cdot)$ is a symmetric kernel function, and the density is also defined through kernel-weighted summation: $\rho_t(\mathbf{x}|\mathbf{p}_t) = \sum_j K(\mathbf{x}-\mathbf{p}_{t,j})$. Visual particles at timestep $t$ are characterized by their attributes $\{\mathbf{x}_{t}, \mathbf{c}_{t}, \mathbf{s}_t, \mathbf{o}_t, \mathbf{r}_t\}$, representing position, color, scale, opacity, and orientation, respectively. Similarly, there are $N_t^\text{visual}$ visual particles at $t$. They are rendered through rasterization following the 3D Gaussian Splatting~\citep{kerbl20233d}.

\paragraph{Physical Constraints from Simulation.}
We generate physical constraints for both reconstruction and prediction through fluid simulation. The core idea is to use simulation to provide a physically plausible guess of the physical particles $\mathbf{p}_t^{\textrm{sim}}$, which, together with incompressibility, creates a physics loss to help solve for the velocity. We will only optimize $\mathbf{p}_t$ as it exclusively represents the velocity field\footnote{The velocity field $\mathbf{V}_t$ is a function of $\mathbf{u}_t$ and $\mathbf{p}_t$, and $\mathbf{u}_t = (\mathbf{p}_t-\mathbf{p}_{t-1})/\Delta t$ is also a function of $\mathbf{p}_t$ given fixed $\mathbf{p}_{t-1}$.} and density field. The simulation step can be written as:
$\mathbf{p}_t^{\textrm{sim}} = \texttt{Sim}(\mathbf{u}_{t-1}, \mathbf{p}_{t-1})$,
where we use $\mathbf{p}_t^{\textrm{sim}}$ for computing physics loss and for initializing $\mathbf{p}_t$.
The physics loss consists of two terms:
\begin{equation}
\mathcal{L}_{\textrm{physics}} = \lambda_\text{sim}\mathcal{L}_{\textrm{sim}} + \mathcal{L}_{\textrm{incomp}},
\end{equation}
where $\mathcal{L}_{\textrm{sim}} = \lVert\mathbf{p}_t-\mathbf{p}_t^{\textrm{sim}}\rVert_2^2$ encourages the physical fields to be consistent with the simulation solution, and $\lambda_\text{sim}$ denotes a weight. And we have the incompressibility loss:
\begin{align} \nonumber
\mathcal{L}_{\textrm{incomp}} &= \sum_j(\frac{\rho_t(\mathbf{p}_{t,j})}{\rho_0} - 1)^2 + \lambda_\text{next}\sum_j(\frac{\rho_{t+1}(\mathbf{p}_{t+1,j})}{\rho_0}-1)^2 
\\&+ \lambda_\text{v-incomp}\sum_{i\neq j}(\max(0,\sigma-d_{t,ij}))^2,
\end{align}
where $\rho_0$ denotes the constant environmental air density, and $\mathbf{p}_{t+1}=\texttt{Sim}(\mathbf{u}_t, \mathbf{p}_t)$ denotes the physical particle positions at the next timestep by differentiable simulation. The first two terms encourage incompressibility around the physical particle positions at both current and next timesteps. The third term encourages incompressibility around the visual particles by maintaining a minimal distance $\sigma$ between each pair of them at the current timestep $d_{t,ij}=\lVert\mathbf{x}_{t,i}-\mathbf{x}_{t,j}\rVert_2^2$, where the current positions
\begin{equation}
\mathbf{x}_t = \texttt{Adv}(\mathbf{V}_t, \mathbf{x}_{t-1})
\label{eqn:advect}
\end{equation}
are estimated via advecting $\mathbf{x}_{t-1}$ using the current velocity field $\mathbf{V}_t$. We leave the details of the simulation operator \texttt{Sim} and advection operator \texttt{Adv} in the supplementary material.

\subsection{Generative Reconstruction and Prediction}

\paragraph{Reconstruction.}
Given multi-view videos $\mathcal{V}_{c=0}^C$ with their camera poses $\{\pi_{c}\}_{c=0}^C$, we reconstruct fluid velocity (represented by $\mathbf{p}_t$) and appearance (represented by $\{\mathbf{x}_{t}, \mathbf{c}_{t}, \mathbf{s}_t, \mathbf{o}_t, \mathbf{r}_t\}$) over time $T$. We formulate this as a sequential optimization from $t=1$ to $t=T$, fixing all quantities at $t-1$ when solving for time $t$ (we leave $t=0$ in supplementary material.) Our overall loss function is:
\begin{equation}
\mathcal{L} = \mathcal{L}_{\textrm{physics}} + \mathcal{L}_{\textrm{visual}} + \mathcal{L}_{\textrm{reg}},
\end{equation}
where the visual loss $\mathcal{L}_{\textrm{visual}} = \sum_{c=0}^C \mathcal{L}_1(I_t^c, I'^c_t) + \mathcal{L}_\text{SSIM}(I_t^c, I'^c_t)$ measures differences between reference frames $I^c_t$ and rendered images $I'^c_t=\texttt{Render}(\pi_{c}, \mathbf{x}_{t}, \mathbf{c}_{t}, \mathbf{s}_t, \mathbf{o}_t, \mathbf{r}_t)$ with the image difference losses $\mathcal{L}_1$ and $\mathcal{L}_\text{SSIM}$ as in 3D Gaussian Splatting~\citep{kerbl20233d}.
Here $\mathbf{x}_t$ is also obtained via Eq.~\ref{eqn:advect}. The regularization term $\mathcal{L}_{\textrm{reg}}$ encourages temporal consistency of $\{\mathbf{c}_{t}, \mathbf{s}_t, \mathbf{o}_t, \mathbf{r}_t\}$ with $\{\mathbf{c}_{t-1}, \mathbf{s}_{t-1}, \mathbf{o}_{t-1}, \mathbf{r}_{t-1}\}$ (details in supplementary material).

This sequential optimization is challenging due to potential error accumulation and the complex entanglement between physics and appearance. Therefore, we decompose it into two stages. First, we solve for dynamics while fixing appearance attributes:
\begin{equation}
\min_{\mathbf{p}_t} \mathcal{L}_{\textrm{physics}} + \mathcal{L}_{\textrm{visual}}, \quad t=1,\ldots,T.
\end{equation}
Then, with $\mathbf{p}_t$ fixed, we use Eq.~\ref{eqn:advect} to get $\mathbf{x}_t$ and optimize other appearance attributes:
\begin{equation}
\min_{\mathbf{c}_t, \mathbf{s}_t, \mathbf{o}_t, \mathbf{r}_t} \mathcal{L}_{\textrm{visual}} + \mathcal{L}_{\textrm{reg}}, \quad t=1,\ldots,T.
\end{equation}

\paragraph{Limitations of Pure Physics Simulation in Prediction.}
In prediction, we solve for $\{\mathbf{p}_t, \mathbf{x}_{t}, \mathbf{c}_{t}, \mathbf{s}_t, \mathbf{o}_t, \mathbf{r}_t\}, t=T+1,\cdots,T_\text{target}$ with optionally new interactions (\eg, inserting an object or adding a wind).
While physics simulation can be highly accurate with known initial conditions and complete fluid properties, predicting future states for reconstructed fluids poses significant challenges. Due to inevitable inaccuracies in the reconstructed velocity and appearance fields, and the absence of modeling specific fluid attributes like temperature and viscosity, pure simulation struggles to capture the complex fluid dynamics of the observed fluid. These inaccuracies compound over time, leading to simplified fluid motion that deviates from the actual dynamics. Furthermore, visual characteristics of the observed fluid, such as scattering effects, cannot be faithfully reproduced through simulation of the reconstructed states alone.

\paragraph{Generative Fluid Simulation.}
We address these limitations through generative simulation that combines physics simulation with video generation. For predicting $t > T$, we first simulate rough fluid dynamics by $\mathbf{p}_t^\text{pred}=\texttt{Sim}(\mathbf{u}_{t-1}^\text{pred},\mathbf{p}_{t-1}^\text{pred})$ and $\mathbf{x}_t^\text{pred}=\texttt{Adv}(\mathbf{V}^\text{pred}_t,\mathbf{x}_{t-1}^\text{pred})$ from $t=T+1$ to $t=T_\text{target}$, with $\mathbf{x}_T^\text{pred}=\mathbf{x}_T$ and $\mathbf{p}_T^\text{pred}=\mathbf{p}_T$. Then,
we render rough multi-view videos $\{(\hat{I}_{T+1}^c,\cdots,\hat{I}^c_{T_\text{target}})\}_{c=0}^C$ where $\hat{I}_t^c = \texttt{Render}(\pi_{c}, \mathbf{x}_{t}, \mathbf{c}_0, \mathbf{s}_0, \mathbf{o}_0, \mathbf{r}_0), t>T$, where $\mathbf{c}_0, \mathbf{s}_0, \mathbf{o}_0, \mathbf{r}_0$ are constant initialization values. These rough videos capture basic fluid dynamics but lack detailed dynamics or visual realism. We then refine these videos using our video refinement model as in Eq.~\ref{eqn:refinement} to obtain reference prediction videos $\{(I_{T+1}^c,\cdots,I^c_{T_\text{target}})\}_{c=0}^C$ and apply the reconstruction algorithm described above to solve for $t=T+1,\cdots,T_\text{target}$. We leave interaction simulation details in the supplementary material. We also summarize the reconstruction algorithm and the prediction algorithm in the supplementary material.

%% file: fig_text/framework.tex
\begin{figure*}
    \centering
    \includegraphics[width=0.9\textwidth]{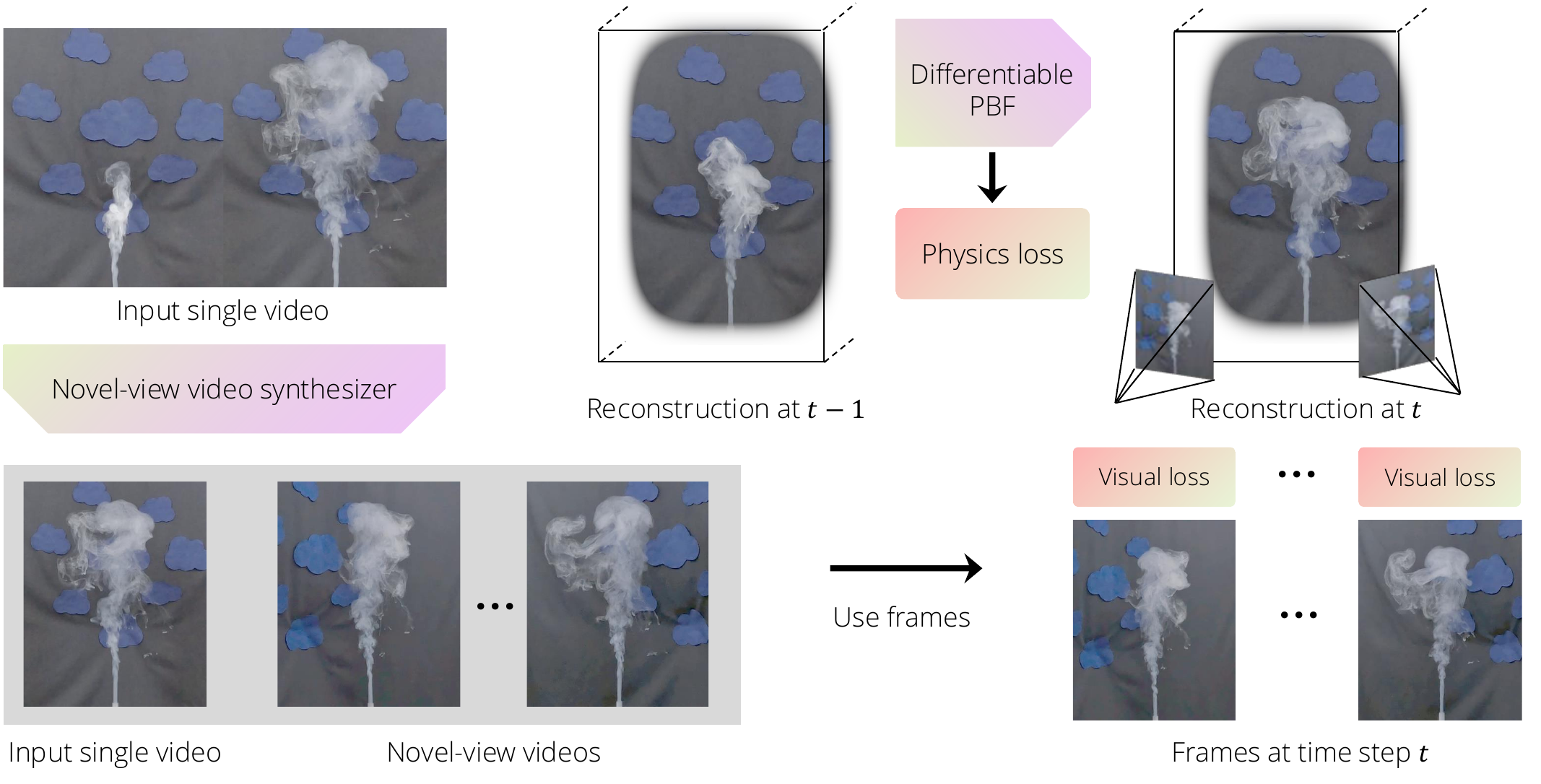}
    \vspace{-4mm}
    \caption{{\bf \model in reconstruction.} From a single video, we synthesize multiple novel-view videos as references for 3D fluid reconstruction. We then sequentially optimize the visual-physical particle fluid representations over time, using the multi-view video frames to compute the visual loss and the physics constraints to compute the physics loss. Our reconstruction output is the 3D fluid appearance and velocity fields over all input frames.
    }
    \label{fig:framework}
    \vspace{-5mm}
\end{figure*}

%% file: sections/4_exp.tex
\section{Experiments}

\input{fig_text/nvs_scalar}
\input{fig_text/nvs}
\input{tab_text/scalar_nums}

\noindent\textbf{Datasets.}~We use the widely-adopted ScalarFlow dataset~\cite{eckert2019scalarflow} which contains $104$ real plume scenes with five synchronized videos for each scene. Yet, all scenes in ScalarFlow do not have textured backgrounds or fluid interaction with other objects. Thus, we collect two real datasets for evaluating fluid reconstruction and prediction in more challenging setups. We leave more details in the supplementary material.

\vspace{0.1cm}
\noindent
\underline{\newdatasetbg} 
includes $120$ scenes. Each scene has a jet of smoke spurting from an ejector in front of a textured background. The ejector creates a burst of fast rising smoke that yields more vortical details compared to ScalarFlow scenes, making this dataset much more challenging. Each scene has $5$ synchronized multi-view videos where the cameras are placed along a horizontal arc of approximately $120\degree$. For each scene, we keep $6$s of the fluid event with FPS $30$ (\ie, $180$ frames). %

\vspace{0.1cm}
\noindent
\underline{\newdatasetball} includes $120$ scenes. Compared to \newdatasetbg, \newdatasetball inserts a shiny ball above the smoke ejector, creating fluid-solid interactions.

\myparagraph{Baselines.}~We compare \model with two representative state-of-the-art approaches on 3D fluid reconstruction:  
Physics-Informed Neural Fluid (PINF)~\cite{chu2022physics} and HyFluid~\cite{yu2024inferring}. In addition, we include a state-of-the-art 4D dynamic reconstruction model SpacetimeGaussians (STG)~\cite{li2024spacetime}.

\myparagraph{Task settings.} For each scene in each dataset, we use the video from the middle camera as input, and the other $4$ videos as ground truth. For 3D fluid reconstruction, we evaluate the visual appearance via novel view synthesis, and we evaluate the velocity via re-simulation (\ie, recreating the fluid dynamics by progressively advecting reconstructed fluid at the initial timestep). For these two tasks, we use the first $120$ frames. For 3D fluid prediction, we focus on future prediction and interaction simulation. To create ground truth for future prediction, we use the first $120$ frames for reconstruction and predict the next $60$ frames for our dataset. %

\myparagraph{Metrics.}~For novel view synthesis and future prediction, we compute PSNR, SSIM~\cite{wang2004image} and LPIPS~\cite{zhang2018perceptual} against the groundtruth frames. For re-simulation, we also compute the divergence of the velocity fields $\nabla\cdot\mathbf{V}$ averaged over time and 3D dense grid points to measure the incompressibility and the quality of the reconstructed velocity field.

\myparagraph{Implementation details.}~We train a Zero123~\cite{zero123} model as our frame-wise novel view synthesis model $g$, which generates frames at a resolution of $256 {\times} 256$. 
We use CogVideoX-5b~\cite{yang2024cogvideox} as our generative video refinement model $v$, which has a time window of $T'{=}49$ frames.
It generates videos at a resolution of $720 {\times} 480$. 
We use the public pretrained model and fine-tune it with LoRA~\cite{hu2021lora} on each of the fluid datasets. 
We use CogVLM2~\cite{wang2023cogvlm, hong2024cogvlm2} to generate captions for our datasets and the ScalarFlow~\cite{eckert2019scalarflow}. 
We set $\lambda_{\text{SDEdit}}{=}0.5$ for refining frames during reconstruction and $\lambda_{\text{SDEdit}}{=}0.75$ during prediction.
For our dataset, we used $100$ scenes as fine-tuning training samples, while for ScalarFlow~\cite{eckert2019scalarflow}, we used $94$ scenes as training samples. We held out $20$ scenes and $10$ scenes, respectively, for evaluation. %
We set the loss weights to $\lambda_\text{sim} = 0.1$, $\lambda_\text{next} = 0.1$, and $\lambda_\text{v-incomp} = 0.1$ for all experiments.
For more details, please refer to supplementary material.

\input{fig_text/future_pred}
\input{tab_text/blue_nums}
\input{tab_text/red_nums}
\subsection{Comparison to baselines}
\label{sec:exp_nvs}

\paragraph{Novel view synthesis.}

We showcase the novel view synthesis results on the ScalarFlow dataset~\citep{eckert2019scalarflow} in Fig.~\ref{fig:nvs_scalar} and the results on our new datasets in Fig.~\ref{fig:nvs_ours}. We observe that PINF~\citep{chu2022physics} and HyFluid~\citep{yu2024inferring}, both designed for multi-view fluid reconstruction using physics losses and neural rendering, fail to synthesize reasonable novel views, as they cannot disambiguate the thickness of the fluid. In contrast, \model allows plausible novel view synthesis on all three datasets. We show quantitative metrics in Tab.~\ref{tab:scalar_nums}, Tab.~\ref{tab:blue_cloud_nums}, and Tab.~\ref{tab:red_ball_nums}, where \model also outperforms all baselines in quantitative measurements.
Note that even for scenes with textured backgrounds and scenes containing fluid-object interaction, \model achieves reasonable 3D fluid reconstruction (Fig.~\ref{fig:nvs_ours}).

\input{fig_text/interaction_wind}
\input{fig_text/interaction_object}

\input{fig_text/ablation}
\input{tab_text/ablation}

\paragraph{Future prediction.}
We showcase input-view future prediction results on the \newdatasetbg in Fig.~\ref{fig:future_pred_iv}. We compared our method with PINF~\citep{chu2022physics} and HyFluid~\citep{yu2024inferring}, noting that their future predictions often mimic the last observed input frame due to an inadequate 3D velocity field, leading to physically inaccurate fluid advection. The single-view input also limits appearance information, resulting in lower-quality predictions. As shown in Fig.~\ref{fig:future_pred_nv}, future predictions from novel viewpoints reveal that other methods fail to deliver satisfactory results. Tab.~\ref{tab:scalar_nums}, Tab.~\ref{tab:blue_cloud_nums}, and Tab.~\ref{tab:red_ball_nums} further demonstrate that while these methods perform reasonably well in the input view, our approach consistently outperforms them across all views.

\paragraph{Re-simulation.}
Tab.~\ref{tab:scalar_nums}, Tab.~\ref{tab:blue_cloud_nums}, and Tab.~\ref{tab:red_ball_nums} present quantitative evaluations of the appearance consistency and physical correctness of the reconstructed velocity fields. Our method \model, which reconstructs fluid particle positions and velocity fields at each time step under physical constraints, achieves lower divergence of velocity field and fully reproduces the reconstruction results. In contrast, other methods show significantly poorer performance. %

\subsection{Interaction Simulation}\label{sec:counter_sim}
To showcase the superiority and robustness of our PBF-based differentiable simulator combined with generative models, we present scenarios involving interactions with external force or additional object. We use \textit{the same input video} for all methods. After reconstruction, we simulate adding wind or a ball to the reconstructed scenes. Our explicit PBD-based~\cite{muller2007position} representation allows direct application of forces or object interaction constraints to particle positions. In contrast, implicit NeRF-based~\cite{nerf} methods like PINF~\citep{chu2022physics} and HyFluid~\citep{yu2024inferring} must reconstruct entire volumes and solve higher-order PDEs, making their process less stable, which is much difficult comparing to our model. Thus, we compare only with SpacetimeGaussians~\cite{li2024spacetime} in this context. While no ground truth videos are available with and without the extra force or rigid body, qualitative results Fig.~\ref{fig:interaction_wind} and Fig.~\ref{fig:interaction_object} show that our method produces more realistic outcomes.

\subsection{Ablation studies}

We conduct ablation study experiments on the \newdatasetbg dataset to evaluate the core components of our proposed \model.

\paragraph{Video generation.} We evaluate three variants: Firstly, we remove the novel-view video synthesizer from \model, denoted as ``w/o NVS''. Secondly, we remove the generative video refinement, denoted as ``w/o GVR''. Lastly, we remove the long video generation support (\ie, we use only the unconditional video refinement diffusion model and refine each $49$-frame segment at a time), denoted as ``w/o LVG''. We show the novel view synthesis results in the left panel of Tab.~\ref{tab:ablation}.
From Tab.~\ref{tab:ablation} we observe that the performance degrades drastically when we remove the novel-view video synthesizer, as it provides the generative priors for a reasonable reconstruction especially for the empty air space and the background region. This can be further observed in Fig.~\ref{fig:ablation}. The noisy floaters in mid-air and the distorted background textures are due to the lack of reconstruction reference from different viewpoints. The generative video refinement and long video generation are also important as they prevent jittering. This can be clearly observed when viewing the video results (attached in the supplementary material). They only provide mild improvements in terms of novel view synthesis metrics, as these metrics do not measure temporal consistency.

\paragraph{Physics constraints.} We evaluate three variants. We remove the physics loss and denote the first variant as ``w/o $\mathcal{L}_{\textrm{physics}}$''. We remove the incompressibility loss and denote the second variant as ``w/o $\mathcal{L}_{\textrm{incomp}}$''. We remove the simulation supervision and denote the third variant as ``w/o $\mathcal{L}_{\textrm{sim}}$''. We show the re-simulation results in the right panel of Tab.~\ref{tab:ablation}.
We conducted re-simulation experiments to evaluate the impact of physical constraints on the reconstructed velocity fields. Without these constraints, image loss dominates, producing velocity fields that lack physical accuracy. As shown in Tab.~\ref{tab:ablation}, these incorrect velocity fields greatly compromise the re-simulation results. Furthermore, Fig.~\ref{fig:ablation} illustrates that the absence of physical constraints results in inaccurate particle trajectories, leading to noticeable artifacts.

%% file: fig_text/nvs_scalar.tex
\begin{figure}[t]
    \centering
    \includegraphics[width=\linewidth]{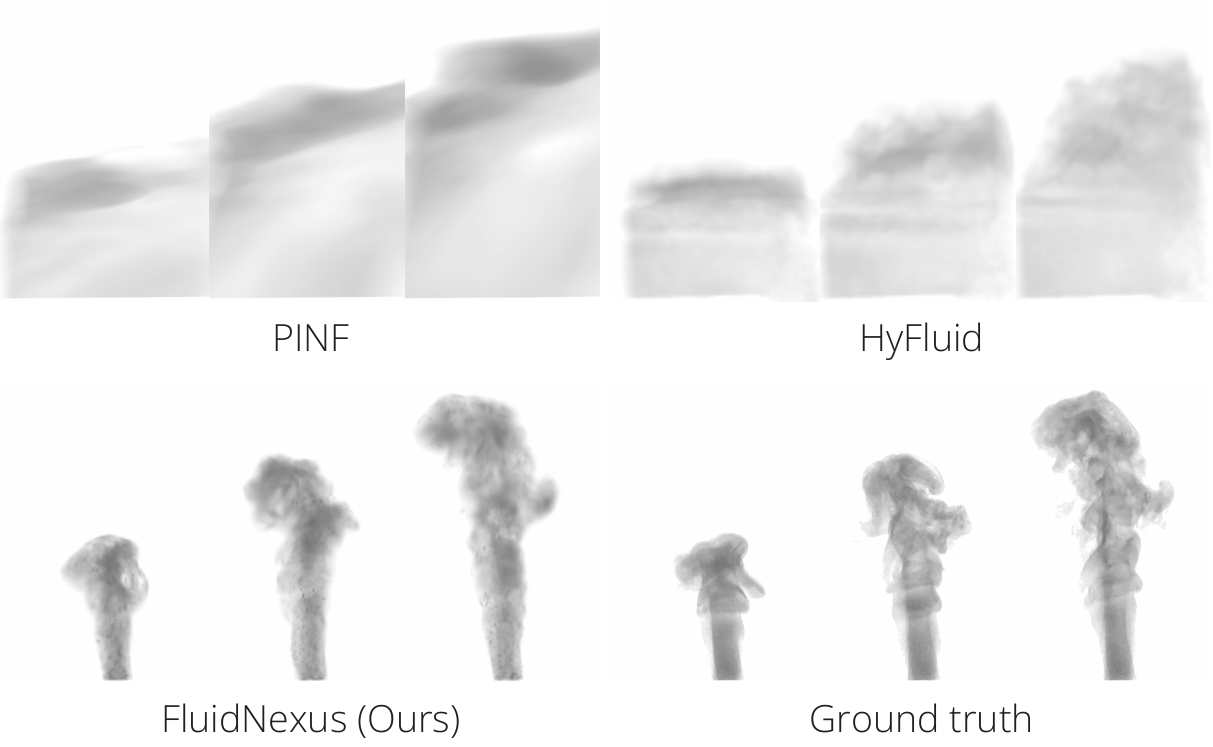}
    \vspace{-5.5mm}
    \caption{Novel view synthesis on ScalarFlow~\cite{eckert2019scalarflow}.
    }
    \label{fig:nvs_scalar}
    \vspace{-4.5mm}
\end{figure}

%% file: fig_text/nvs.tex
\begin{figure*}[t]
    \centering
    \includegraphics[width=0.95\linewidth]{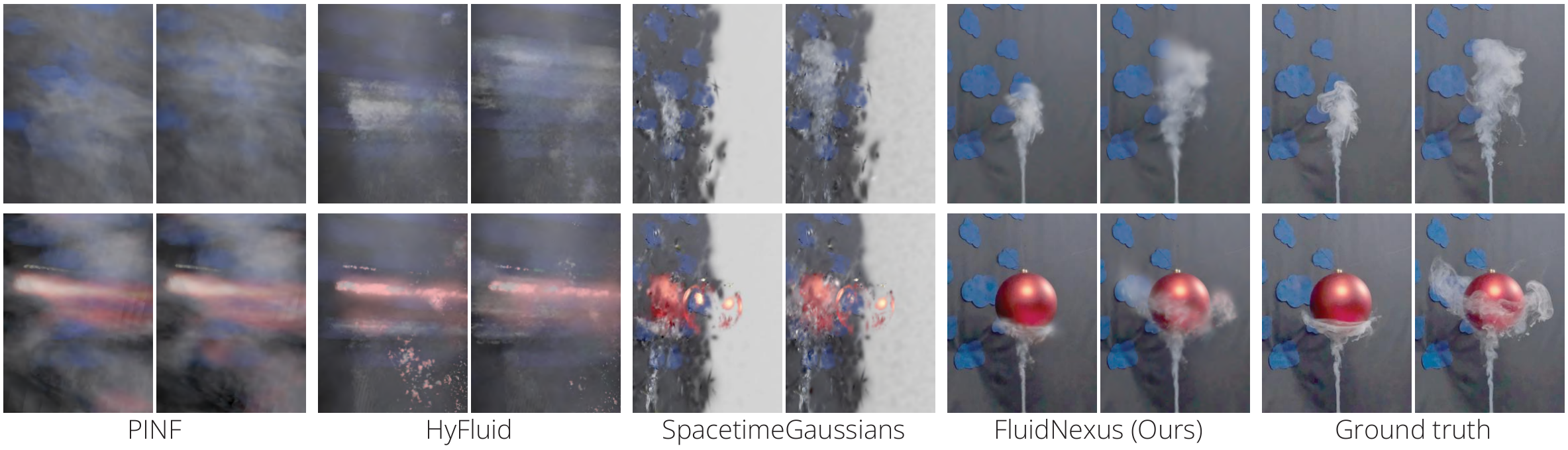}
    \vspace{-3mm}
    \caption{Qualitative results of novel view synthesis on our collected datasets.}
    \label{fig:nvs_ours}
    \vspace{-1.5mm}
\end{figure*}

%% file: tab_text/scalar_nums.tex
\begin{table*}
    [t]
    \centering
    \resizebox{0.96\linewidth}{!}{%
    \begin{tabular}{l|ccc|ccc|ccc|cccc}
        \toprule \multirow{2}{*}{Model}      & \multicolumn{3}{c|}{Novel View Synthesis} & \multicolumn{3}{c|}{Input View Future Prediction} & \multicolumn{3}{c|}{Novel Views Future Prediction} & \multicolumn{4}{c}{Re-simulation} \\
                                             & PSNR $\uparrow$                           & SSIM $\uparrow$                                   & LPIPS $\downarrow$                                 & PSNR $\uparrow$                  & SSIM $\uparrow$   & LPIPS $\downarrow$ & PSNR $\uparrow$  & SSIM $\uparrow$   & LPIPS $\downarrow$ & PSNR $\uparrow$  & SSIM $\uparrow$   & LPIPS $\downarrow$ & $\nabla \cdot \mathbf{V}\downarrow$ \\
        \midrule
        PINF ~\cite{chu2022physics}          & 22.68                                     & 0.7597                                            & 0.1926                                             & 20.48                            & 0.6689            & 0.2737             & 20.66            & 0.6709            & 0.2704             & 22.16            & 0.7409            & 0.1970             & 0.0297                              \\
        HyFluid ~\cite{yu2024inferring}      & 22.23                                     & 0.7645                                            & 0.2275                                             & 26.84                            & 0.9072            & 0.1776             & 20.29            & 0.6280            & 0.3418             & 22.26            & 0.7643            & 0.2272             & 0.0619                              \\
        STG ~\cite{li2024spacetime}          & 19.85                                     & 0.7063                                            & 0.2790                                             & 21.79                            & 0.8759            & 0.2142             & 18.51            & 0.7011            & 0.3697             & 19.73            & 0.7033            & 0.2880             & 0.0973                              \\
        \model (Ours)                        & $\mathbf{32.45}$                          & $\mathbf{0.9544}$                                 & $\mathbf{0.1299}$                                  & $\mathbf{28.51}$                 & $\mathbf{0.9159}$ & $\mathbf{0.1754}$  & $\mathbf{26.83}$ & $\mathbf{0.8952}$ & $\mathbf{0.2052}$  & $\mathbf{32.44}$ & $\mathbf{0.9543}$ & $\mathbf{0.1299}$  & $\mathbf{0.0126}$                   \\
        \bottomrule
    \end{tabular}}
    \vspace{-2mm}
    \caption{Quantitative results on ScalarFlow~\cite{eckert2019scalarflow}.}
    \label{tab:scalar_nums}
    \vspace{-3.5mm}
\end{table*}

%% file: fig_text/future_pred.tex
\begin{figure}[t]
    \centering
    \includegraphics[width=0.9\linewidth]{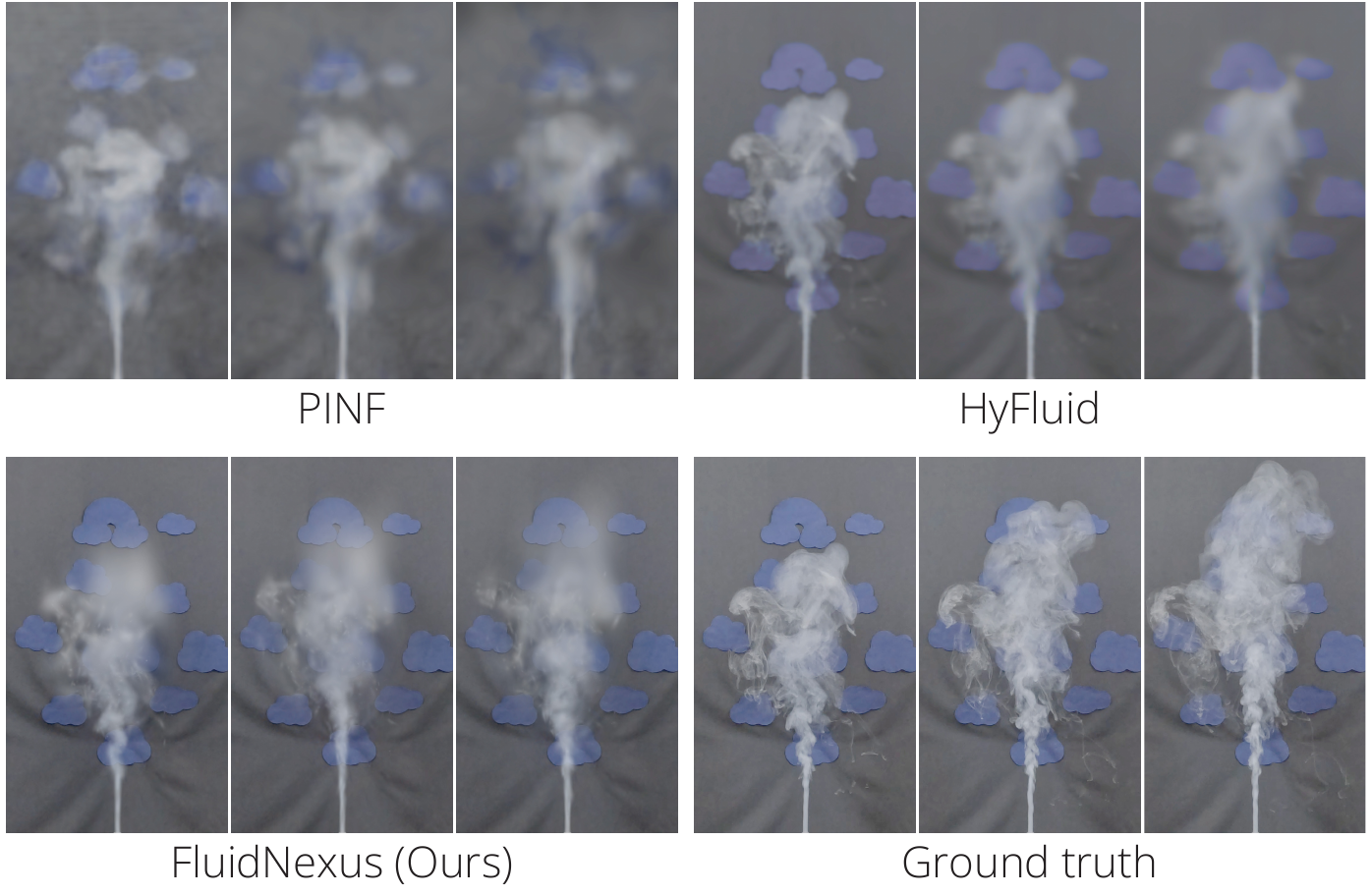}
    \vspace{-3mm}
    \caption{Qualitative results of future prediction on input view.}
    \label{fig:future_pred_iv}
    \vspace{-3mm}
\end{figure}

\begin{figure}[t]
    \centering
    \includegraphics[width=0.9\linewidth]{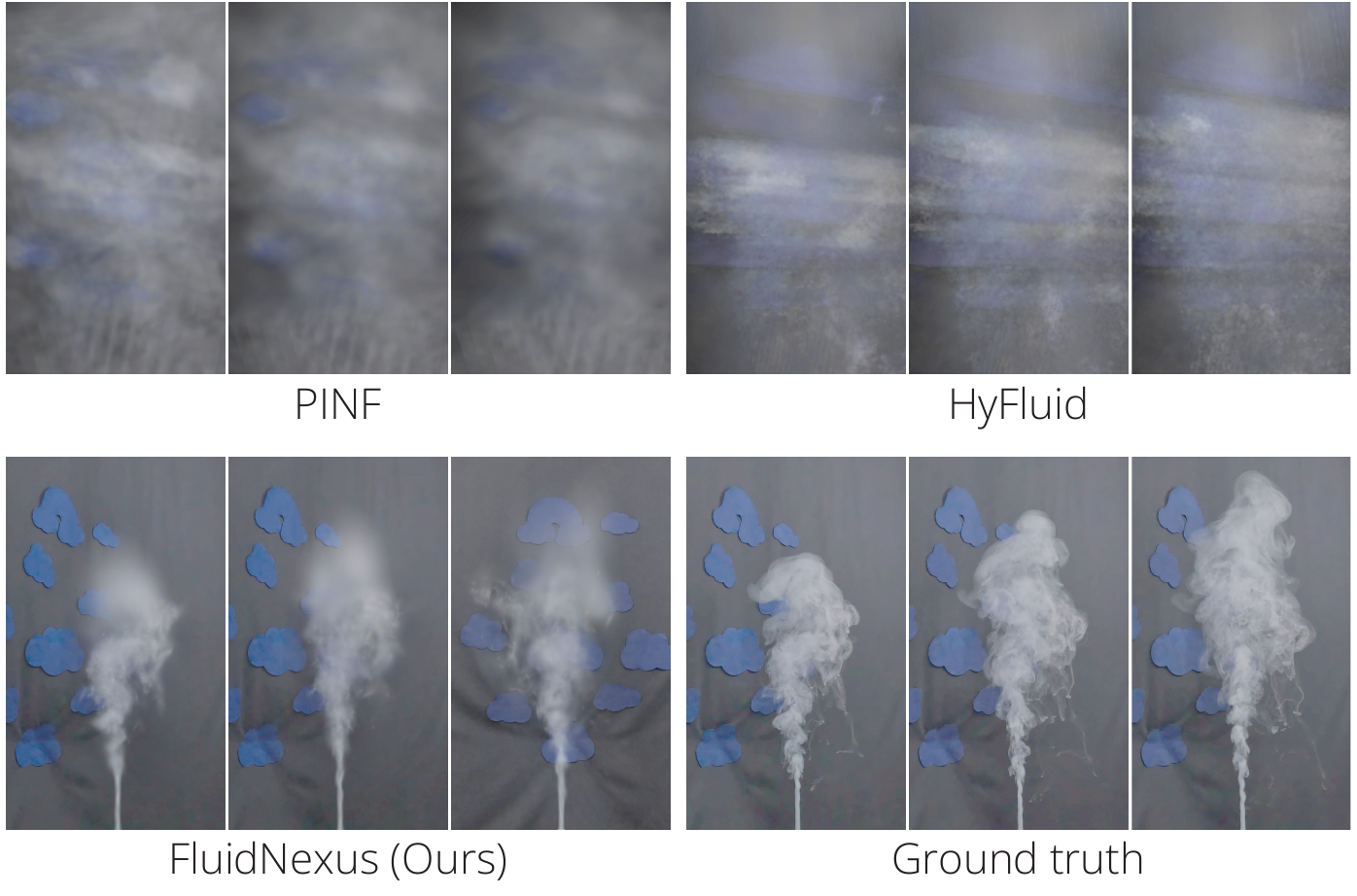}
    \vspace{-3mm}
    \caption{Qualitative results of future prediction on novel view.}
    \label{fig:future_pred_nv}
    \vspace{-6mm}
\end{figure}

%% file: tab_text/blue_nums.tex
\begin{table*}
    [t]
    \vspace{-3mm}
    \centering
    \resizebox{0.96\linewidth}{!}{%
    \begin{tabular}{l|ccc|ccc|ccc|cccc}
        \toprule \multirow{2}{*}{Model} & \multicolumn{3}{c|}{Novel View Synthesis} & \multicolumn{3}{c|}{Input View Future Prediction} & \multicolumn{3}{c|}{Novel Views Future Prediction} & \multicolumn{4}{c}{Re-simulation} \\
                                        & PSNR $\uparrow$                           & SSIM $\uparrow$                                   & LPIPS $\downarrow$                                 & PSNR $\uparrow$                  & SSIM $\uparrow$   & LPIPS $\downarrow$ & PSNR $\uparrow$  & SSIM $\uparrow$   & LPIPS $\downarrow$ & PSNR $\uparrow$  & SSIM $\uparrow$   & LPIPS $\downarrow$ & $\nabla \cdot \mathbf{V}\downarrow$ \\
        \midrule
        PINF ~\cite{chu2022physics}     & 22.40                                     & 0.8002                                            & 0.5089                                             & 26.48                            & 0.7299            & 0.2418             & 22.66            & 0.8234            & 0.5931             & 21.97            & 0.7992            & 0.5029             & 0.0451                              \\
        HyFluid ~\cite{yu2024inferring} & 22.64                                     & 0.7948                                            & 0.4764                                             & 21.14                            & 0.7044            & 0.5417             & 21.95            & 0.8334            & 0.6013             & 22.34            & 0.7615            & 0.4937             & 0.0573                              \\
        STG ~\cite{li2024spacetime}     & 19.94                                     & 0.6875                                            & 0.3673                                             & 23.94                            & 0.8408            & 0.2639             & 18.34            & 0.6325            & 0.4116             & 19.48            & 0.6867            & 0.4818             & 0.0323                              \\
        \model (Ours)                   & $\mathbf{30.62}$                          & $\mathbf{0.9209}$                                 & $\mathbf{0.1707}$                                  & $\mathbf{27.79}$                 & $\mathbf{0.8747}$ & $\mathbf{0.2337}$  & $\mathbf{25.74}$ & $\mathbf{0.8609}$ & $\mathbf{0.2675}$  & $\mathbf{30.61}$ & $\mathbf{0.9208}$ & $\mathbf{0.1707}$  & $\mathbf{0.0246}$                   \\
        \bottomrule
    \end{tabular}}
    \vspace{-2mm}
    \caption{Quantitative results on our \newdatasetbg.}
    \label{tab:blue_cloud_nums}
    \vspace{-2mm}
\end{table*}

%% file: tab_text/red_nums.tex
\begin{table*}
    [t]
    \centering
    \resizebox{0.96\linewidth}{!}{%
    \begin{tabular}{l|ccc|ccc|ccc|cccc}
        \toprule \multirow{2}{*}{Model} & \multicolumn{3}{c|}{Novel View Synthesis} & \multicolumn{3}{c|}{Input View Future Prediction} & \multicolumn{3}{c|}{Novel Views Future Prediction} & \multicolumn{4}{c}{Re-simulation} \\
                                        & PSNR $\uparrow$                           & SSIM $\uparrow$                                   & LPIPS $\downarrow$                                 & PSNR $\uparrow$                  & SSIM $\uparrow$   & LPIPS $\downarrow$ & PSNR $\uparrow$  & SSIM $\uparrow$   & LPIPS $\downarrow$ & PSNR $\uparrow$  & SSIM $\uparrow$   & LPIPS $\downarrow$ & $\nabla \cdot \mathbf{V}\downarrow$ \\
        \midrule
        PINF ~\cite{chu2022physics}     & 20.50                                     & 0.7556                                            & 0.4611                                             & 24.70                            & 0.8162            & 0.5079             & 18.26            & 0.6199            & 0.4458             & 20.05            & 0.7111            & 0.5127             & 0.0441                              \\
        HyFluid ~\cite{yu2024inferring} & 20.71                                     & 0.7251                                            & 0.4978                                             & 20.93                            & 0.7005            & 0.5019             & 18.28            & 0.7239            & 0.4380             & 20.60            & 0.7152            & 0.5192             & 0.0518                              \\
        STG ~\cite{li2024spacetime}     & 18.70                                     & 0.6793                                            & 0.3866                                             & 25.80                            & 0.8629            & 0.2251             & 17.11            & 0.6156            & 0.4131             & 18.46            & 0.6819            & 0.3843             & 0.0531                              \\
        \model (Ours)                   & $\mathbf{29.89}$                          & $\mathbf{0.9107}$                                 & $\mathbf{0.1773}$                                  & $\mathbf{27.70}$                 & $\mathbf{0.8733}$ & $\mathbf{0.2019}$  & $\mathbf{24.82}$ & $\mathbf{0.8431}$ & $\mathbf{0.2607}$  & $\mathbf{29.88}$ & $\mathbf{0.9101}$ & $\mathbf{0.1729}$  & $\mathbf{0.0280}$                   \\
        \bottomrule
    \end{tabular}}
    \vspace{-2mm}
    \caption{Quantitative results on our \newdatasetball.}
    \label{tab:red_ball_nums}
    \vspace{-6mm}
\end{table*}

%% file: fig_text/interaction_wind.tex
\begin{figure}[t]
    \centering
    \includegraphics[width=\linewidth]{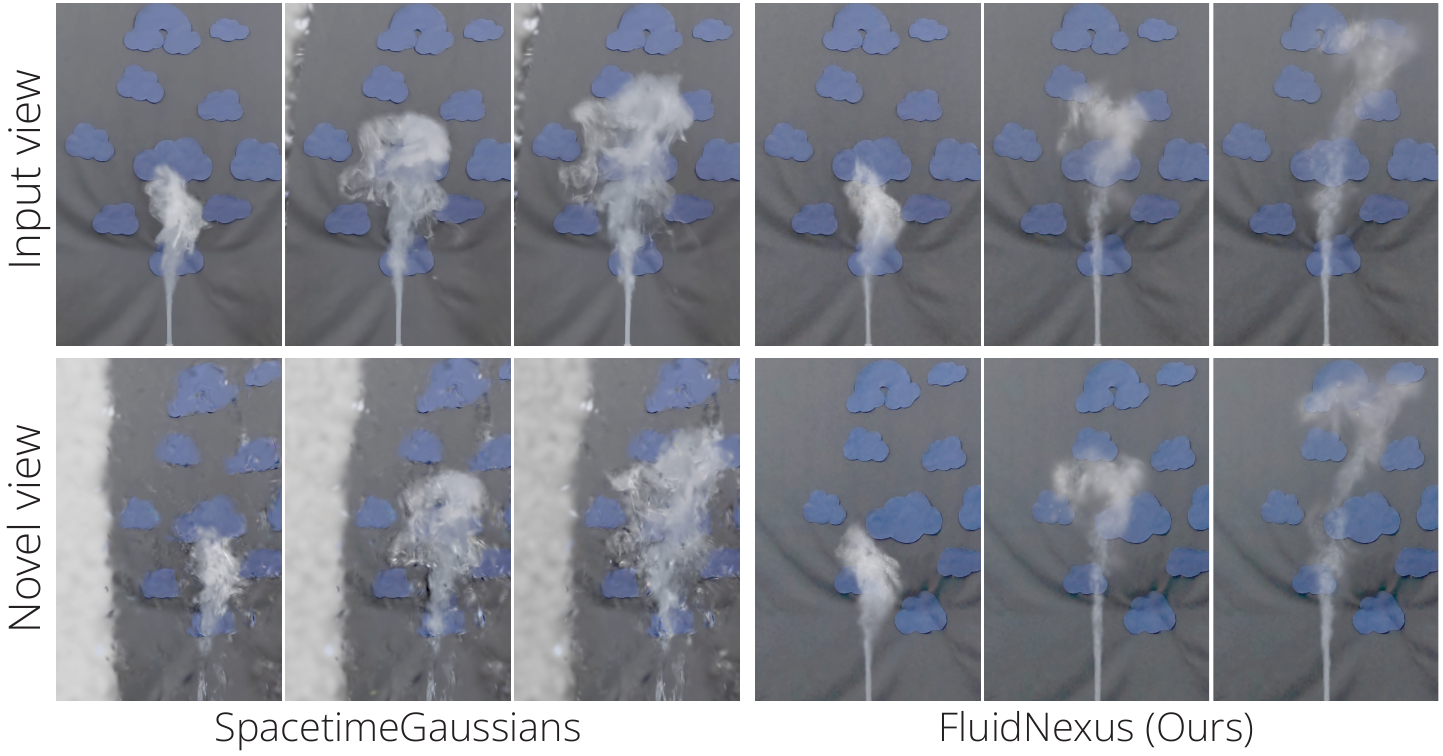}
    \vspace{-6mm}
    \caption{Qualitative results of wind-fluid interactive simulation results rendered at the input view or the novel view.}
    \label{fig:interaction_wind}
    \vspace{-3mm}
\end{figure}

%% file: fig_text/interaction_object.tex
\begin{figure}[t]
    \centering
    \includegraphics[width=\linewidth]{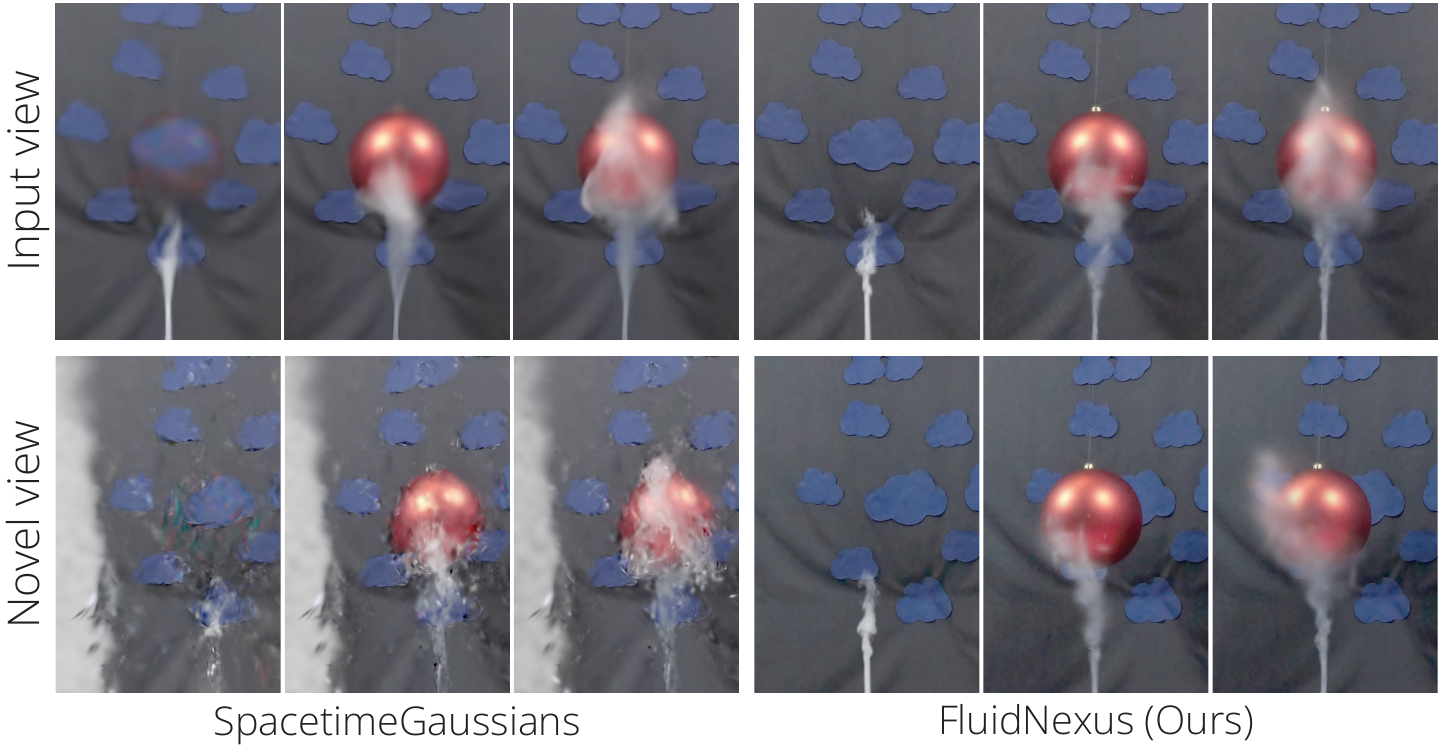}
    \vspace{-6mm}
    \caption{Qualitative results of object-fluid interactive simulation results rendered at the input view or the novel view. }
    \label{fig:interaction_object}
    \vspace{-3mm}
\end{figure}

%% file: fig_text/ablation.tex
\begin{figure*}
    \centering
    \includegraphics[width=0.99\linewidth]{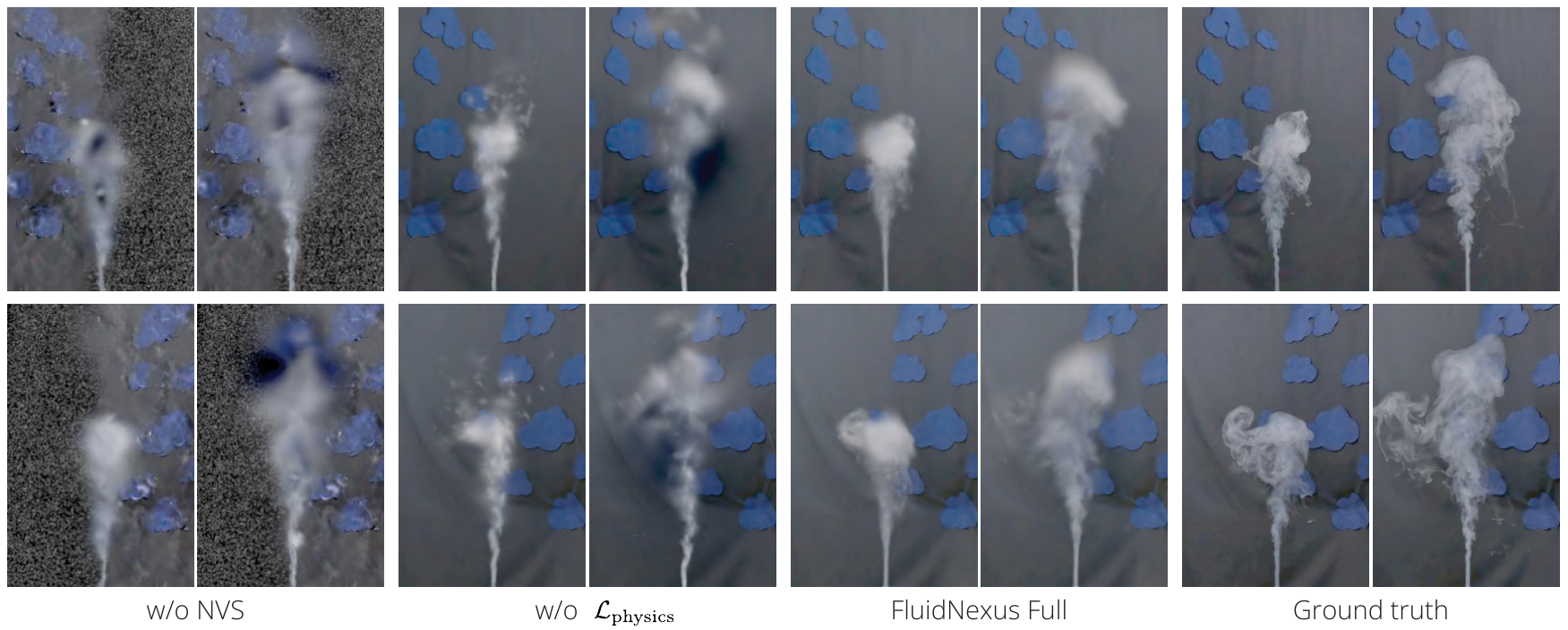}
    \vspace{-2.5mm}
    \caption{Ablation studies of \model. Results in the top row are from a novel view, and results in the bottom row are from another novel view. We leave a visual comparison of all variants in the supplementary material.
    }
    \vspace{-3mm}
    \label{fig:ablation}
\end{figure*}

%% file: tab_text/ablation.tex
\begin{table}[t]
    \centering
    \resizebox{\linewidth}{!}{%
    \begin{tabular}{lccc|lccc}
        \toprule \multirow{2}{*}{Settings} & \multicolumn{3}{c|}{Novel View Synthesis} & \multirow{2}{*}{Settings} & \multicolumn{3}{c}{Re-simulation} \\
                                           & PSNR $\uparrow$                           & SSIM $\uparrow$           & LPIPS $\downarrow$               &                                      & PSNR $\uparrow$  & SSIM $\uparrow$   & LPIPS $\downarrow$ \\
        \midrule
        w/o NVS                            & 21.88                                     & 0.5909                    & 0.6117                           & w/o $\mathcal{L}_{\textrm{physics}}$ & 28.00            & 0.9016            & 0.1913             \\
        w/o GVR                            & 29.12                                     & 0.9060                    & 0.1904                           & w/o $\mathcal{L}_{\textrm{incomp}}$  & 28.76            & 0.9148            & 0.1854             \\
        w/o LVG                            & 30.14                                     & 0.9196                    & 0.1819                           & w/o $\mathcal{L}_{\textrm{sim}}$     & 29.70            & 0.9175            & 0.1905             \\
        \model                             & $\mathbf{30.62}$                          & $\mathbf{0.9209}$         & $\mathbf{0.1707}$                & \model                               & $\mathbf{30.61}$ & $\mathbf{0.9208}$ & $\mathbf{0.1707}$  \\
        \bottomrule
    \end{tabular}}
    \vspace{-1mm}
    \caption{Ablation studies on \newdatasetbg.}
    \label{tab:ablation}
    \vspace{-4mm}
\end{table}

%% file: sections/5_conclusion.tex
\section{Conclusion}
We presented \model, a novel framework that enables 3D fluid reconstruction and prediction from a single video by bridging video generation with physics simulation. Through extensive experiments on two new challenging fluid datasets, we show that video generation provides significant support for future prediction tasks. Our work facilitates further research on single-video fluid analysis.

\paragraph{Limitations.} One limitation of \model is that the reconstruction results on challenging datasets, including \newdatasetbg and \newdatasetball, are blurry, missing fine fluid details. The other major limitation is that \model does not generalize to different backgrounds or different objects, as the novel view video synthesizer is trained on limited scenes. Collecting large-scale multi-view fluid video datasets, either synthetic or real, can be a promising direction to improve generalization.

%% file: sections/6_ack.tex
\myparagraph{Acknowledgments.} 
The work was in part supported by ONR MURI N00014-24-1-2575, ONR MURI N00014-22-1-2740, and NSF RI \#2211258 and \#2338203.

%% file: sections_supp/0_overview.tex
\section{Overview}
In this supplementary material, we elaborate on the details of our approach (\ref{sec:supp_tech}), datasets (\ref{sec:supp_dataset}), and experiments (\ref{sec:supp_exp}). We also include results on an in-the-wild scene from HyFluid~\citep{yu2024inferring} and results on using multiple views rather than a single view (\ref{sec:supp_results}). We further discuss ablation study results (\ref{sec:supp_ablation}).
We compile video results in our project website \url{https://yuegao.me/FluidNexus}. We strongly encourage the readers to review the video results first.

%% file: sections_supp/1_tech_details.tex
\section{Technical Details}\label{sec:supp_tech}

\paragraph{Position-based fluid simulation.} Our physics simulation is based on position-based fluid (PBF)~\citep{macklin2013position,macklin2014unified} which extends position-based dynamics (PBD)~\citep{muller2007position,macklin2014unified}. PBD provides a simple and flexiable particle-based simulation framework based on solving position constraints. In the following, we briefly review the PBF simulation. We refer the reader to Macklin et.al.~\citep{muller2007position,macklin2014unified} for more details. In short,
PBD uses a set of particles to represent the scene and each particle consists of its position, mass, and velocity. PBD solves a system of non-linear equality and inequality constraints for the position correction $\boldsymbol{\Delta} \mathbf{p}$ such that physical constraints are met:
\begin{equation}
\begin{aligned}
& c_i(\mathbf{p}+\boldsymbol{\Delta} \mathbf{p})=0, \quad i=1, \ldots, n, \\
& c_j(\mathbf{p}+\boldsymbol{\Delta} \mathbf{p}) \geq 0, \quad j=1, \ldots, n,
\end{aligned}
\end{equation}
where $\mathbf{p}=\left[\mathbf{p}_1, \mathbf{p}_2, \ldots, \mathbf{p}_{N}\right]^T$ denotes the vector of particle positions.
Constraints are solved sequentially using the linearization of $C$ around $\mathbf{p}$, and the position change $\boldsymbol{\Delta} \mathbf{p}$, is restricted to lie along the constraint gradient:
\begin{equation}
c_i(\mathbf{p}+\boldsymbol{\Delta} \mathbf{p}) \approx c_i(\mathbf{p})+\nabla c_i(\mathbf{p}) \boldsymbol{\Delta} \mathbf{p}=0,
\end{equation}
and the particle velocity is then given by $\mathbf{v}=\frac{\boldsymbol{\Delta}\mathbf{p}}{\Delta t}$. We assume that all physical particles have a mass of $1$ to simplify the formulation.

The constraint for fluid (incompressibility) is given by
\begin{equation}
    c_\text{fluid}(\mathbf{p}_1, \ldots, \mathbf{p}_{N_\text{physical}})=\frac{\rho_i}{\rho_0}-1 = 0,
\end{equation}
where the fluid density is estimated by $\rho_i=\sum_{j} K(\mathbf{p}_i-\mathbf{p}_j)$ and $K$ is a kernel function. We use the cubic Poly6 kernel~\citep{muller2003particle}. We also utilize the drag force as an external force to model the effect of fast moving air interacting with the surrounding environment:
\begin{align}
\label{eq:force}
\mathbf{f}_i = - k (\mathbf{v}_i-\mathbf{v}_\text{env})\max(0,1 - \frac{\rho_i}{\rho_o}),
\end{align}
where $\mathbf{v}_\text{env}$ is the environmental velocity at the particle position and is set to  $\mathbf{0}$ to model still air, and $k>0$ is the drag force coefficient.

\paragraph{Initialization.}
To initialize the fluid simulation, we follow Macklin et.al.~\citep{macklin2014unified} to add a source region and run $N_\text{stable}$ stabilization simulation steps at $t=0$ (\ie, before the reconstruction starts). We seed physical particles and visual particles within the source region at each timestep. After the stabilization steps, we obtain the initial physical particles $\{\mathbf{p}_0,\mathbf{u}_0\}$ and visual particle positions $\{\mathbf{x}_0\}$. All other visual particle attributes are initialized as constants $\{\mathbf{c}_0, \mathbf{s}_0, \mathbf{o}_0, \mathbf{r}_0\}$ for all timesteps.

\paragraph{Simulation and advection operators.}
The simulation operator for the physical particle position $\mathbf{p}_t^{\textrm{sim}} = \texttt{Sim}(\mathbf{u}_{t-1}, \mathbf{p}_{t-1})$ consists of three steps: (1) We generate a guess of particle velocities $\hat{\mathbf{u}}_{t}=\mathbf{u}_{t-1}+\Delta t \cdot \alpha \cdot \mathbf{g} + \Delta t \cdot \mathbf{f}$ where $\mathbf{g}$ denotes the gravity and $\alpha<0$ denotes the buoyancy coefficient. $\mathbf{f}$ denotes an optional external force. (2) We generate a guess of physical particle positions $\hat{\mathbf{p}}_t = \mathbf{p}_{t-1}+\Delta t \cdot \hat{\mathbf{u}}_t$. (3) We obtain the simulation result $\mathbf{p}^\text{sim}_t$ by solving $c_\text{fluid}(\hat{\mathbf{p}}_t+\boldsymbol{\Delta}\mathbf{p}_{t})=0$. 

For the advection operator $\texttt{Adv}$, we use a simple forward Euler integrator, \ie, $\texttt{Adv}(\mathbf{V}, \mathbf{x}) = \mathbf{x} + \Delta t \cdot \mathbf{V}(\mathbf{x})$, while advanced advection schemes can also be used \eg, BFECC~\citep{selle2008unconditionally}.

\paragraph{Regularization term.}
To encourage temporal smoothness we add a regularization term $\mathcal{L}_\text{reg}$:
\begin{align}
    \nonumber
\mathcal{L}_\text{reg} &= \lambda_c\|\mathbf{c}_t - \mathbf{c}_{t-1}\|_2^2 + \lambda_s\|\mathbf{s}_t - \mathbf{s}_{t-1}\|_2^2 \\ &+ \lambda_o\|\mathbf{o}_t - \mathbf{o}_{t-1}\|_2^2 + \lambda_r\|\mathbf{r}_t - \mathbf{r}_{t-1}\|_2^2 + \mathcal{L}_\text{aniso},
\end{align}
where $\lambda_c$, $\lambda_s$, $\lambda_o$, and $\lambda_r$ are weighting coefficients for the color, scale, opacity, and orientation terms respectively. The L2 terms help maintain smooth transitions in the visual attributes. We also adopt the anisotropic loss $\mathcal{L}_\text{aniso}$ from Xie et.al.~\citep{xie2024physgaussian} to prevent overly skinny visual particles.

\input{fig_text/flowchart}

\input{algo_text/reconstruction}

\input{algo_text/prediction}
\paragraph{Algorithms and flowchart.}
We summarize the fluid reconstruction algorithm in Alg.~\ref{algo:reconstruction}, and the fluid prediction algorithm in Alg.~\ref{algo:prediction}.
We further show a flowchart in Figure~\ref{fig:flowchart} to visualize the entire pipeline of \model.

%% file: fig_text/flowchart.tex
\begin{figure*}
    \centering
    \includegraphics[width=1\textwidth]{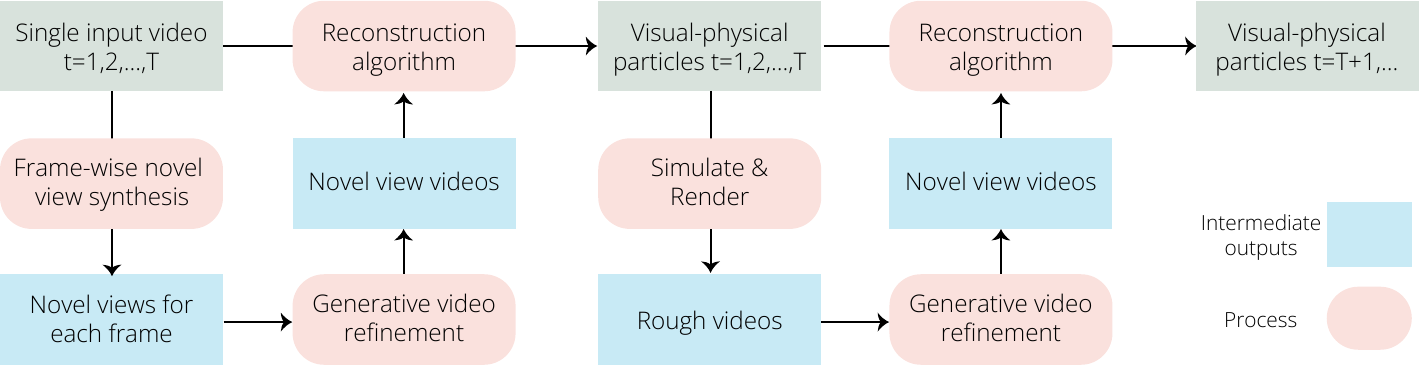}
    \caption{{\bf Flowchart of \model.} The reconstruction algorithm is shown in Alg.~\ref{algo:reconstruction}. On the left is the reconstruction process which takes a single video as input and produces the visual-physical particles within the video duration. On the right is the prediction process which takes the reconstructed visual-physical particles as input and produces the particles in future timesteps. 
    }
    \label{fig:flowchart}
\end{figure*}

%% file: algo_text/reconstruction.tex
\begin{algorithm}[H]
    \caption{\model: Reconstruction}
    \label{algo:reconstruction}

    \begin{algorithmic}

    \State \textbf{Input:} Multi-view videos $\{\mathcal{V}^c\}_{c=0}^C$, camera poses $\{\pi_{c}\}_{c=0}^C$

    \State \textbf{Initialize:} Physical particles $\{\mathbf{p}_0, \mathbf{u}_0\}$ and visual particles $\{\mathbf{x}_0, \mathbf{c}_0, \mathbf{s}_0, \mathbf{o}_0, \mathbf{r}_0\}$ through stabilization

    \Statex

    \State \textbf{Physical particle reconstruction for $t=1$ to $T$:}
    \begin{itemize}
    \setlength{\itemindent}{0.5em}
        \item[\textbf{1.}] Simulate physical guess $\mathbf{p}_t^{\textrm{sim}} = \texttt{Sim}(\mathbf{u}_{t-1}, \mathbf{p}_{t-1})$
        \item[\textbf{2.}] Optimize $\mathbf{p}_t$ by minimizing:
            \begin{itemize}
                \item Simulation loss: $\mathcal{L}_{\textrm{sim}} = \|\mathbf{p}_t-\mathbf{p}_t^{\textrm{sim}}\|_2^2$
                \item Incompressibility loss: $\mathcal{L}_{\textrm{incomp}}$ on current and next timesteps
                \item Visual loss: $\mathcal{L}_{\textrm{visual}}$ across all views
            \end{itemize}
        \item[\textbf{3.}] Update velocity $\mathbf{u}_t = (\mathbf{p}_t-\mathbf{p}_{t-1})/\Delta t$
    \end{itemize}
    
    \Statex
    
    \State \textbf{Visual particle reconstruction for $t=1$ to $T$:}
    \begin{itemize}
    \setlength{\itemindent}{0.5em}
        \item[\textbf{1.}] Compute visual positions $\mathbf{x}_t = \texttt{Adv}(\mathbf{V}_t, \mathbf{x}_{t-1})$
        \item[\textbf{2.}] Optimize $\{\mathbf{c}_t, \mathbf{s}_t, \mathbf{o}_t, \mathbf{r}_t\}$ by minimizing:
            \begin{itemize}
                \item Visual loss: $\mathcal{L}_{\textrm{visual}}$ across all views
                \item Temporal regularization: $\mathcal{L}_{\textrm{reg}}$ with previous timestep
            \end{itemize}
    \end{itemize}
    
    \State \textbf{Output:} Reconstructed fluid velocity by $\{\mathbf{p}_t, \mathbf{u}_t\}_{t=1}^T$ and appearance by $\{\mathbf{x}_t, \mathbf{c}_t, \mathbf{s}_t, \mathbf{o}_t, \mathbf{r}_t\}_{t=1}^T$
    \end{algorithmic}
\end{algorithm}

%% file: algo_text/prediction.tex
\begin{algorithm}[H]
    \caption{\model: Prediction}
    \label{algo:prediction}
    
    \begin{algorithmic}
    
    \State \textbf{Input:} Reconstructed states $\{\mathbf{p}_T, \mathbf{u}_T\}$ and $\{\mathbf{x}_T, \mathbf{c}_T, \mathbf{s}_T, \mathbf{o}_T, \mathbf{r}_T\}$, camera poses $\{\pi_{c}\}_{c=0}^C$
    
    \Statex
    
    \State \textbf{Initial simulation for $t=T+1$ to} $T_\text{target}$\textbf{:}
    \begin{itemize}
    \setlength{\itemindent}{0.5em}
        \item[\textbf{1.}] Simulate physical particles: $\mathbf{p}_t^\text{pred} = \texttt{Sim}(\mathbf{u}_{t-1}^\text{pred}, \mathbf{p}_{t-1}^\text{pred})$
        \item[\textbf{2.}] Advect visual particles: $\mathbf{x}_t^\text{pred} = \texttt{Adv}(\mathbf{V}_t^\text{pred}, \mathbf{x}_{t-1}^\text{pred})$
        \item[\textbf{3.}] Render rough multi-view frames: $\hat{I}_t^c = \texttt{Render}(\pi_{c}, \mathbf{x}_t^\text{pred}, \mathbf{c}_0, \mathbf{s}_0, \mathbf{o}_0, \mathbf{r}_0)$
    \end{itemize}
    
    \Statex
    
    \State \textbf{Video refinement:} Generate reference videos $\{(I_{T+1}^c,\cdots,I^c_{T_\text{target}})\}_{c=0}^C$ using the generative video refinement model $v$ on each rough video $(\hat{I}_{T+1}^c,\cdots,\hat{I}^c_{T_\text{target}})$
    
    \Statex
    
    \State \textbf{Physical particle reconstruction for $t=T+1$ to} $T_\text{target}$\textbf{:}
    \begin{itemize}
    \setlength{\itemindent}{0.5em}
        \item[\textbf{1.}] Simulate physical guess $\mathbf{p}_t^{\textrm{sim}} = \texttt{Sim}(\mathbf{u}_{t-1}, \mathbf{p}_{t-1})$
        \item[\textbf{2.}] Optimize $\mathbf{p}_t$ by minimizing:
            \begin{itemize}
                \item Simulation loss: $\mathcal{L}_{\textrm{sim}} = \|\mathbf{p}_t-\mathbf{p}_t^{\textrm{sim}}\|_2^2$
                \item Incompressibility loss: $\mathcal{L}_{\textrm{incomp}}$ on current and next timesteps
                \item Visual loss: $\mathcal{L}_{\textrm{visual}}$ across all views
            \end{itemize}
        \item[\textbf{3.}] Update velocity $\mathbf{u}_t = (\mathbf{p}_t-\mathbf{p}_{t-1})/\Delta t$
    \end{itemize}
    
    \Statex
    
    \State \textbf{Visual particle reconstruction for $t=T+1$ to} $T_\text{target}$\textbf{:}
    \begin{itemize}
    \setlength{\itemindent}{0.5em}
        \item[\textbf{1.}] Compute visual positions $\mathbf{x}_t = \texttt{Adv}(\mathbf{V}_t, \mathbf{x}_{t-1})$
        \item[\textbf{2.}] Optimize $\{\mathbf{c}_t, \mathbf{s}_t, \mathbf{o}_t, \mathbf{r}_t\}$ by minimizing:
            \begin{itemize}
                \item Visual loss: $\mathcal{L}_{\textrm{visual}}$ across all views
                \item Temporal regularization: $\mathcal{L}_{\textrm{reg}}$ with previous timestep
            \end{itemize}
    \end{itemize}
    
    \State \textbf{Output:} Predicted fluid dynamics $\{\mathbf{p}_t, \mathbf{u}_t\}_{t=T+1}^{T_\text{target}}$ and appearance $\{\mathbf{x}_t, \mathbf{c}_t, \mathbf{s}_t, \mathbf{o}_t, \mathbf{r}_t\}_{t=T+1}^{T_\text{target}}$
    
    \end{algorithmic}
\end{algorithm}

%% file: sections_supp/2_dataset_details.tex
\section{Dataset Details}\label{sec:supp_dataset}
\input{fig_text/setup}
\input{fig_text/raw_data}

We show a photo and an illustration of our dataset capture setup in Figure~\ref{fig:setup}. A black cloth was used as the background, onto which patches of various shapes were attached. To eliminate the influence of ambient light, we employed two adjustable color-temperature light sources. A handheld portable fog generator (Fog Machine Model S) with remote start and stop control created the desired smoke effects. The other objects on the ground were only used to hold down the black cloth to prevent it from moving and are not directly related to our setup. Other objects were used to add textures to help camera calibration.

The setup included six GoPro HERO 12 cameras, five of which were fixed to their locations and used as primary data recording cameras, while the sixth served as a secondary camera to capture multiple images for camera calibration using COLMAP~\cite{schoenberger2016sfm, schoenberger2016mvs}. The GoPro cameras were mounted on tripods and configured to a $5$K ($2988{\times}5312$) resolution with $1.4\times$ magnification. The raw frames captured by the cameras are shown in Figure~\ref{fig:raw_data}. We center-cropped the frames to $1440{\times}2560$ and resized them to $1080{\times}1920$. All experiments and camera calibrations were conducted at a resolution of $1080{\times}1920$. The frame rate was set to $50$ fps to address power-line flicker. During preprocessing, all videos were converted into individual frames, and we sampled the frames with a step size of $2$ for all experiments. The frame rate was set to $30$ fps for presenting videos and conducting experiments. 

Although we used a program to control the start and stop of the cameras, slight mis-synchronization could still occur. To address this, we manually labeled the starting frame of each video to ensure frame synchronization across all viewpoints.

For both of our datasets, we recorded $120$ scenes each. We used $100$ scenes as training data and evaluated the model on the remaining $20$ scenes. Additionally, for the ScalarFlow~\cite{eckert2019scalarflow} dataset, we used $94$ scenes for training and evaluated on the remaining $10$ scenes.

%% file: fig_text/setup.tex
\begin{figure*}
    \centering
    \includegraphics[width=\textwidth]{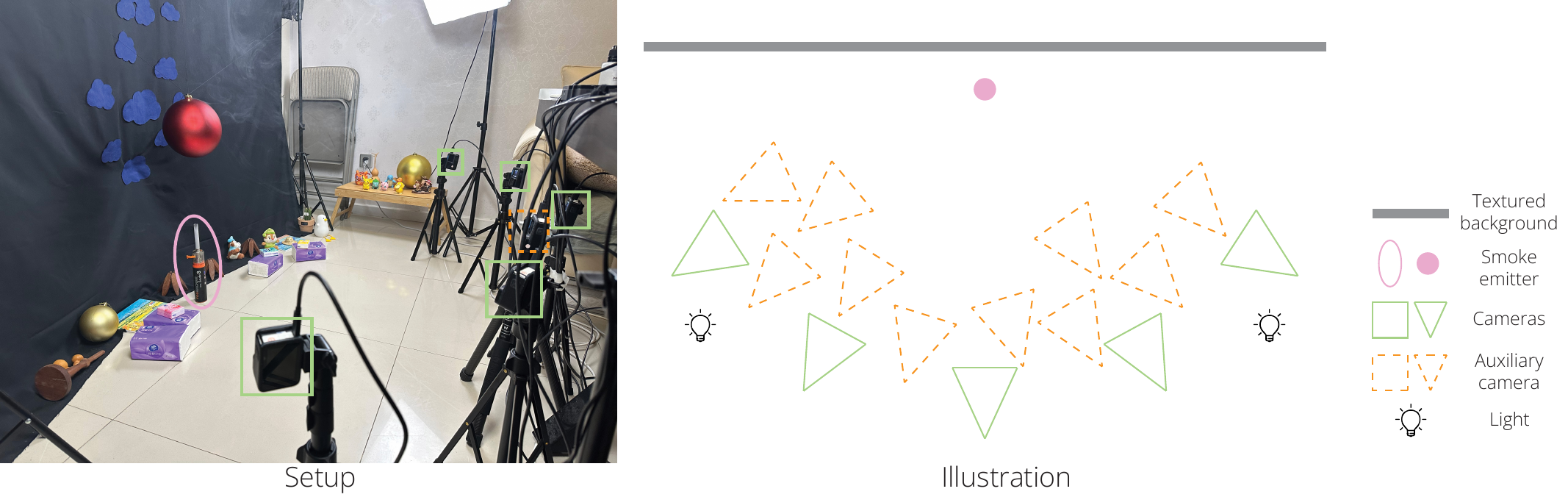}
    \vspace{-5mm}
    \caption{ The configuration of our data capture system.
    }
    \label{fig:setup}
    \vspace{1mm}
\end{figure*}

%% file: fig_text/raw_data.tex
\begin{figure*}
    \centering
    \includegraphics[width=\textwidth]{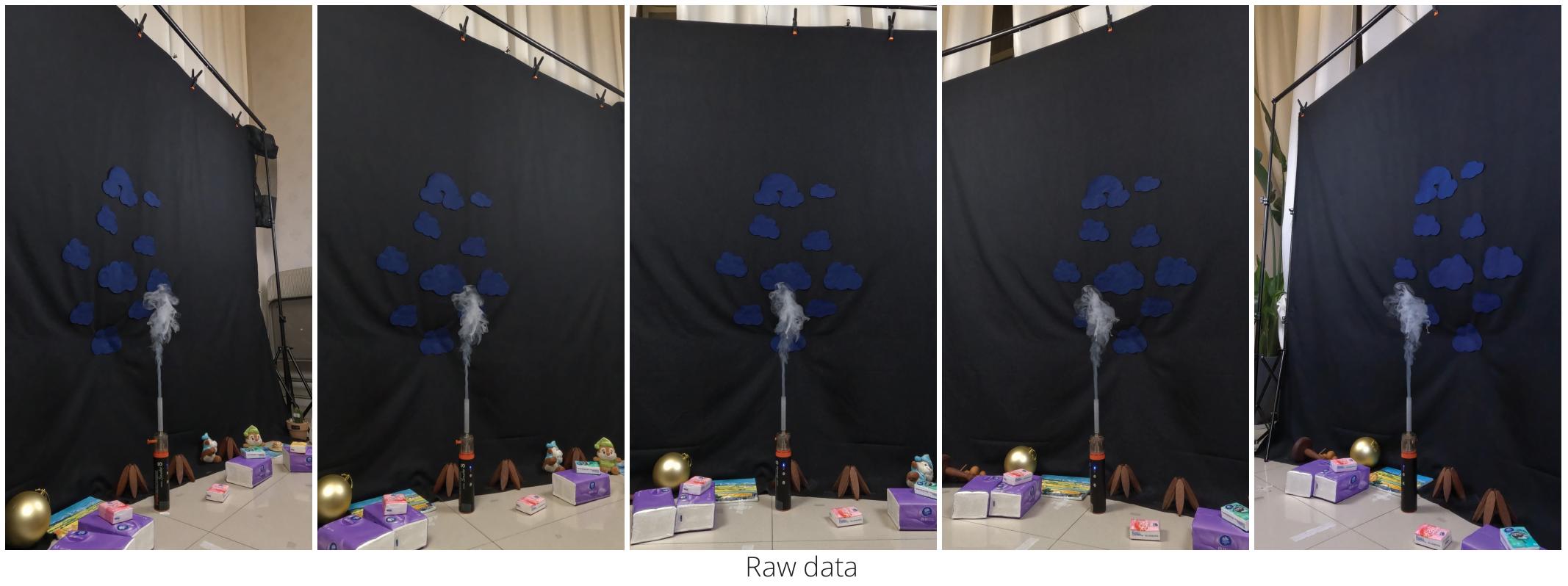}
    \vspace{-5mm}
    \caption{The raw frames captured by our five cameras.
    }
    \label{fig:raw_data}
    \vspace{-3mm}
\end{figure*}

%% file: sections_supp/3_exp_details.tex
\section{Further Implementation Details}\label{sec:supp_exp}

\paragraph{Simulation details.}
In our experiments, we set the timestep $\Delta t=1/30$ seconds. We set the drag force coefficient $k=3$. We set the regularization loss weights $\lambda_c=10$, $\lambda_s=0$, $\lambda_o=8$, $\lambda_r=0.1$. For the ScalarFlow dataset, we set the buoyancy coefficient $\alpha=-3$ and environmental density $\rho_0=2$. For the \newdatasetbg and \newdatasetball datasets, we set $\alpha=-6$ and $\rho_0=1.5$. We set the stabilization simulation steps to $N_\text{stable}=20$. We set the PBF constraint solver iteration count to $10$. This maintains mean simulation speed of $0.04493$ seconds per simulation timestep (amounting to $22.26$ FPS).

\vspace{2mm}
\input{fig_text/supp_zero123}
\paragraph{Frame-wise novel view synthesis model training.}
We illustrate our frame-wise novel view synthesis model $g$ in Fig.~\ref{fig:rebut_zero123}. It is based on an image diffusion model. In particular, it takes an input-view frame and a target-view camera pose as control signals, and it gradually denoises a Gaussian noise to generate the target-view frame. The generated target-view frames are the input to the generative video refinement module. We utilized Zero123~\cite{zero123} as our image diffusion model. We performed full-parameter fine-tuning based on the official implementation and the pre-trained Zero123-XL~\cite{zero123} model, which is trained on the Objaverse-XL~\cite{objaversexl} dataset.  We used a smaller batch size of $92$ while keeping other hyper-parameters (such as the base learning rate) unchanged.

We fine-tune it on three datasets individually. For the ScalarFlow dataset, we fine-tune the model for $15,000$ iterations. For \newdatasetbg and \newdatasetball, we fine-tune it for $50,000$ iterations. We keep all other official settings and use the official implementation of Zero123~\cite{zero123}. For our experiments, we apply square padding and resize it to $256{\times}256$ to match the input and output frame size of the original implementation. We maintain the aspect ratio by cropping the output results and resizing them back to a resolution of $1080{\times}1920$.

\paragraph{Generative video refinement model training.}
For the generative video refinement model $v$, we use CogVideoX~\cite{hong2022cogvideo} as our base model and fine-tune it. We use the official implementation of CogVideoX~\cite{hong2022cogvideo}. We utilized CogVLM2~\cite{wang2023cogvlm} to generate captions for all videos, and fine-tuned the model for $10,000$ iterations across all datasets using the official LoRA~\cite{hu2021lora} implementation included in CogVideoX's official implementation. Similar to training the view synthesis model, we pad and resize our dataset to the resolution required by CogVideoX~\cite{hong2022cogvideo} ($720{\times}480$), and inversely transform its output back to our experimental resolution ($1080{\times}1920$).

\paragraph{Interaction simulation.}
For the counterfactual interaction simulation in Sec. \textcolor{red}{4.2}, we consider two types of interaction: external body force (\eg, the wind) and one-way coupling with rigid body (\eg, the ball). External body force is simply implemented by setting $\mathbf{f}$. Specifically, we set the force $\mathbf{f}$ such that it is exponentially increasing with the physical particle position along the $y$ axis (opposite gravity direction). The one-way coupling is implemented by fixing the rigid object still and use the simplified contact constraints from Macklin et.al.~\cite{macklin2014unified}. In particular, we use 3D Gaussian splatting to reconstruct the ball, which gives particles along the surface of the ball. The constraint for the fluid's physical particles is that if a particle enters the interior of the rigid object, we find the nearest object surface particle and update the physical fluid particle position to the exterior of the rigid body.

%% file: fig_text/supp_zero123.tex
\begin{figure}[t]
    \centering
    \includegraphics[width=0.9\linewidth]{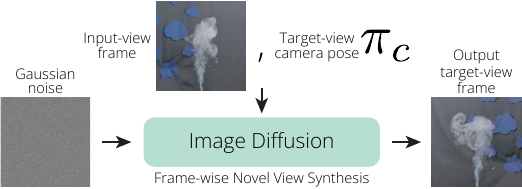}
    \vspace{-1mm}
    \caption{Illustration: Frame-wise Novel View Synthesis module.}
    \label{fig:rebut_zero123}
    \vspace{-2mm}
\end{figure}

%% file: sections_supp/4_more_results.tex
\section{More Results}\label{sec:supp_results}

\paragraph{In-the-wild scene.}
We reconstruct an in-the-wild example from the HyFluid~\cite{yu2024inferring} paper. We use $3$ available input views, as our video synthesizer requires training data. We showcase our re-simulation results in Fig.~\ref{fig:rebut_wild}, and please check our website for video results. Ours significantly outperforms other methods.

\input{fig_text/supp_wild}

\paragraph{Multi-view reconstruction.}
We use $4$ views as input and $1$ holdout view for testing. We show results in Tab.~\ref{tab:rebut_nums} and Fig.~\ref{fig:rebut_nvs}, as well as video results in our website. Our approach significantly outperforms prior methods. It indeed improves performances compared to ours using a single-view input.

\input{tab_text/supp_nums}

\input{fig_text/supp_nvs}

%% file: fig_text/supp_wild.tex
\begin{figure}[t]
    \centering
    \includegraphics[width=0.9\linewidth]{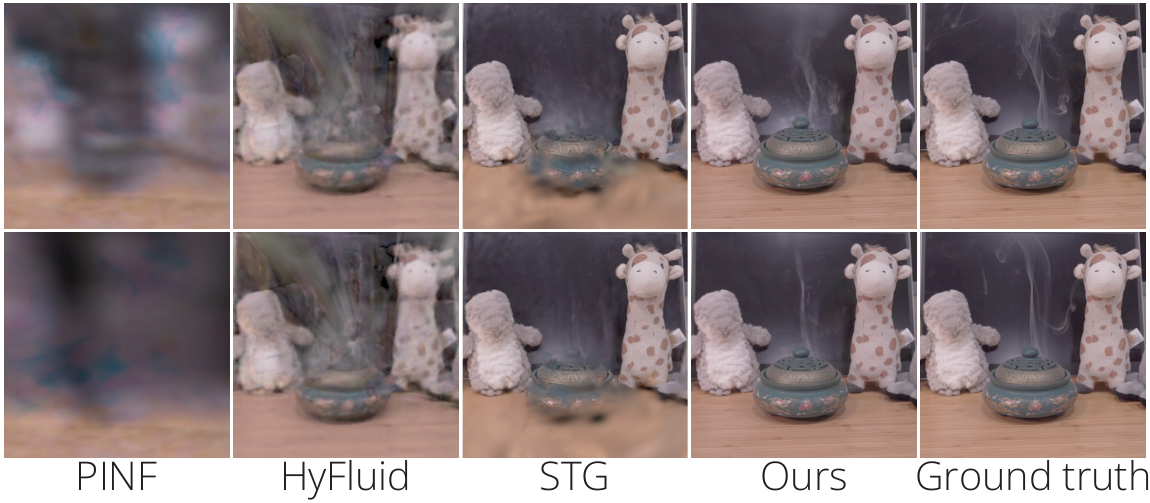}
    \vspace{-2mm}
    \caption{Qualitative results of re-simulation on in-the-wild data.}
    \label{fig:rebut_wild}
    \vspace{-2.6mm}
\end{figure}

%% file: tab_text/supp_nums.tex
\begin{table}[t]
    \centering
    \resizebox{0.989\linewidth}{!}{%
    \begin{tabular}{l|ccc|ccc|ccc}
        \toprule \multirow{2}{*}{Model} & \multicolumn{3}{c|}{Novel View Synthesis} & \multicolumn{3}{c|}{Future Prediction} & \multicolumn{3}{c}{Re-simulation} \\
                                        & PSNR $\uparrow$                           & SSIM $\uparrow$                        & LPIPS $\downarrow$               & PSNR $\uparrow$  & SSIM $\uparrow$   & LPIPS $\downarrow$ & PSNR $\uparrow$  & SSIM $\uparrow$   & LPIPS $\downarrow$ \\
        \midrule PINF                   & 24.45                                     & 0.8568                                 & 0.4973                           & 20.42            & 0.7816            & 0.5883             & 19.44            & 0.6176            & 0.5672             \\
        HyFluid                         & 26.42                                     & 0.8676                                 & 0.4146                           & 22.91            & 0.8181            & 0.6049             & 22.46            & 0.8381            & 0.5623             \\
        STG                             & 19.58                                     & 0.7101                                 & 0.4224                           & 18.48            & 0.6105            & 0.4929             & 19.40            & 0.6098            & 0.5219             \\
        Ours-1                          & 30.43                                     & 0.9212                                 & 0.1812                           & 25.74            & 0.8609            & 0.2675             & 30.42            & 0.9211            & 0.1812             \\
        Ours                            & $\mathbf{31.15}$                          & $\mathbf{0.9216}$                      & $\mathbf{0.1233}$                & $\mathbf{26.84}$ & $\mathbf{0.8654}$ & $\mathbf{0.2382}$   & $\mathbf{31.14}$ & $\mathbf{0.9215}$ & $\mathbf{0.1233}$  \\
        \midrule PINF                   & 21.17                                     & 0.7507                                 & 0.5317                           & 20.08            & 0.7195            & 0.5344             & 18.38            & 0.5618            & 0.5533             \\
        HyFluid                         & 23.92                                     & 0.8327                                 & 0.4346                           & 21.65            & 0.8088            & 0.5907             & 20.29            & 0.8165            & 0.5689             \\
        STG                             & 18.79                                     & 0.7209                                 & 0.3884                           & 18.45            & 0.5982            & 0.5054             & 18.68            & 0.6209            & 0.4883             \\
        Ours-1                          & 28.24                                     & 0.9080                                 & 0.1719                           & 24.82            & 0.8431            & 0.2607             & 28.23            & 0.9079            & 0.1720             \\
        Ours                            & $\mathbf{30.42}$                          & $\mathbf{0.9211}$                      & $\mathbf{0.1121}$                & $\mathbf{26.26}$ & $\mathbf{0.8479}$ & $\mathbf{0.2105}$  & $\mathbf{30.41}$ & $\mathbf{0.9211}$ & $\mathbf{0.1122}$  \\
        \bottomrule
    \end{tabular}}
    \vspace{-1mm}
    \caption{Quantitative results on \newdatasetbg (upper) and \newdatasetball (lower)
    with $4$ input views. ``Ours-1'' denotes the performances using a single view
    as input.}
    \label{tab:rebut_nums}
    \vspace{-2mm}
\end{table}

%% file: fig_text/supp_nvs.tex
\begin{figure}[t]
    \centering
    \includegraphics[width=0.9\linewidth]{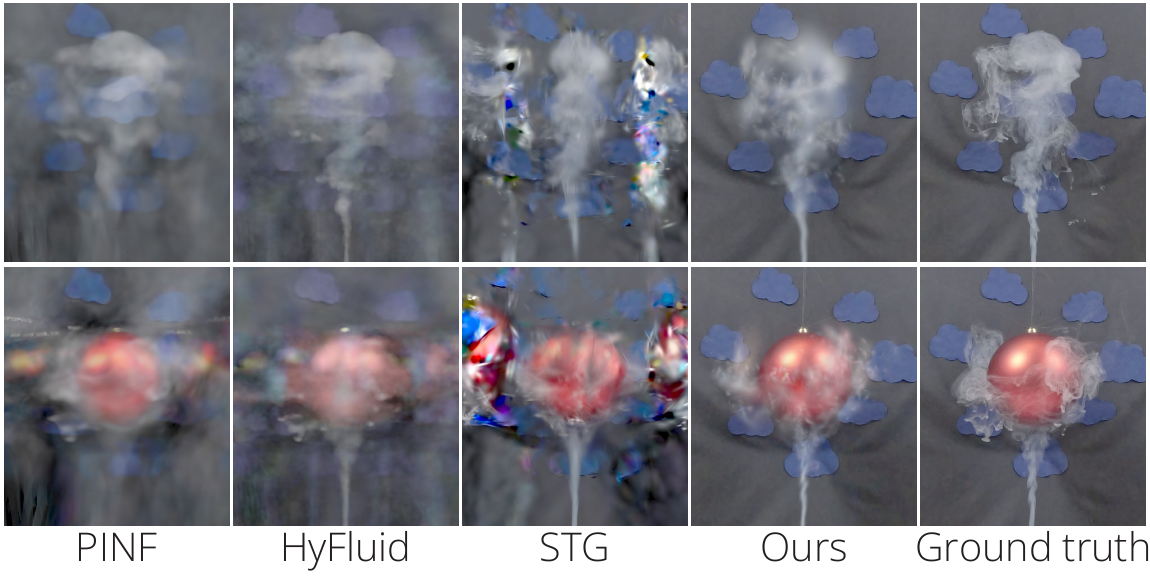}
    \vspace{-2mm}
    \caption{Novel view synthesis results with $4$ input views.}
    \label{fig:rebut_nvs}
    \vspace{-2.6mm}
\end{figure}

%% file: sections_supp/5_ablation.tex
\section{Further Ablation Comparison}\label{sec:supp_ablation}

We provide all variants of the ablation experiments in the video results, showcasing the two tasks of novel view synthesis and re-simulation.

Firstly, in the novel view synthesis task, we observe that when our method excludes the novel-view video synthesizer (``w/o NVS''), the results degrade significantly due to the lack of multi-view constraints. Additionally, when we remove the generative video refinement component (``w/o GVR''), the results include noticeably more jittering artifacts. Furthermore, we can observe abrupt transitions in the video results when the long video generation is removed (``w/o LVG'').

To validate the importance of our physics constraints, we conducted comparisons on the re-simulation task. First, when the physical loss is removed (``w/o $\mathcal{L}_{\textrm{physics}}$''), the video results degrade significantly. This is because the optimized velocity field lacks physical accuracy, leading to implausible particle dynamics during advection in the re-simulation, which results in numerous artifacts. Additionally, when we remove the incompressibility loss (``w/o $\mathcal{L}_{\textrm{incomp}}$''), the visual dynamics look unnatural with mild jittering. When the simulation loss is removed (``w/o $\mathcal{L}_{\textrm{sim}}$''), the visual results also look unnatural with abrupt transitions.